\documentclass{article}

\usepackage{arxiv}
\usepackage[utf8]{inputenc} 
\usepackage[T1]{fontenc}    
\usepackage{url}            
\usepackage{microtype}      %
\usepackage{graphicx}
\usepackage{multirow}
\usepackage{booktabs}
\usepackage{amssymb}
\usepackage{amsmath}
\usepackage[table]{xcolor}
\usepackage{xspace}         

\usepackage[colorlinks=true,linkcolor=blue,citecolor=blue,urlcolor=blue]{hyperref}

\makeatletter
\DeclareRobustCommand\onedot{\futurelet\@let@token\@onedot}
\def\@onedot{\ifx\@let@token.\else.\null\fi\xspace}

\makeatother

\title{SAGE-OR: Semi-supervised Adaptive Scene Graph Generation for Operating Rooms}

\author{
  Brandon Leblanc \\
  The Immersive and Creative Technologies Lab\\
  Concordia University\\
  Montreal, CA \\
  \texttt{brandon.leblanc@mail.concordia.ca} \\
  \And
  Charalambos Poullis \\
  The Immersive and Creative Technologies Lab\\
  Concordia University\\
  Montreal, CA \\
  \texttt{charalambos.poullis@concordia.ca} \\
}

\date{}

\renewcommand{\undertitle}{Accepted at BMVC 2026}
\renewcommand{\headeright}{Accepted at BMVC 2026}

\begin{document}
\maketitle

\makeatletter
\renewcommand{\@title}{SAGE-OR}
\makeatother

\begin{abstract}
Current surgical scene graph generation methods depend on dense multi-modal supervision and specialized hardware (synchronized RGB-D sensors, calibration rigs), making dataset construction expensive and restricting all existing benchmarks to simulated environments. We propose SAGE-OR, a feature-centric framework that replaces the traditional detect-then-reason paradigm with a decoupled representation-reasoning paradigm in which localization is derived from frozen foundation models, encoded implicitly in pre-computed features, and used without any localization supervision, while a lightweight graph transformer performs relational reasoning over cached features. We employ a semi-supervised formulation with general-purpose segmentation prompts to eliminate localization supervision while enabling unsupervised context augmentation through additional prompt-driven entities, such as hands, which are absent from annotations. General-purpose prompts are used to induce near-perfect recall, while precision is delegated to downstream attention-based reasoning, enabling simple adaptation to new entities via prompt-level modification. This design enables a lightweight 15M-parameter graph transformer that trains in 1.4 hours and runs relational inference at ${\sim}$1ms per frame with peak memory under 2GB, suitable for edge hardware used in the operating room; feature extraction runs offline as a separate caching stage (4.27s per frame). On the 4D-OR benchmark, the core model achieves 76\% F1, matching the fully supervised 4D-OR baseline while eliminating all localization annotations, and unsupervised hand augmentation raises this to 86\%, within 4 points of state-of-the-art (SOTA) methods requiring dense multi-modal supervision, providing a practical pathway for adaptation to new surgical settings without annotation other than relationship and class labels. Code and models are publicly available at \href{https://github.com/TheFourthKaramazov/SAGE-OR}{https://github.com/TheFourthKaramazov/SAGE-OR}.
\end{abstract}

\section{Introduction}

Modern surgical procedures are increasingly conducted in complex, dynamic operating room (OR) environments that require intricate coordination between clinicians, patients, and specialized equipment. This complexity is expected to grow as robotic-assisted interventions become more prevalent \cite{ding2025visual, moglia2021systematic, yuan2024advancing, wang2023dynamic}. Effective OR management is therefore essential not only to improve patient outcomes, but also to optimize team performance, enable new surgical technologies, and minimize delays \cite{murali2023latent, ban2024concept}. Scene graph generation (SGG), which represents entities and their interactions as structured graphs in which nodes represent objects and the edges represent corresponding relationships \cite{johnson2015image, tripathi2019compact}, has emerged as an interpretable framework for both external and internal OR perspectives, supporting downstream applications such as workflow recognition, safety monitoring, and context-aware assistance systems \cite{ozsoy2022_4D_OR, ozsoy2024mmor, allan2020robotic, nwoye2023cholectriplet, nwoye2023cholectriplet2022, alhajj2019cataracts, murali2023endoscapes, srivastav2018mvor}. Despite recent advances, widespread deployment of SGG systems in real clinical settings remains impractical due to two key challenges: data scarcity and adaptability \cite{henriques2025decoding}.

First, existing methods are dependent on dense multi-modal supervision that makes dataset construction prohibitively expensive. Current approaches such as LABRAD-OR \cite{ozsoy2023_LABRAD_OR}, S$^2$Former-OR \cite{pei2025sgg}, and TriTemp-OR \cite{guo2024trimodal} require per-frame bounding boxes, human poses, segmentation masks, scene point clouds, object point clouds, and metric depth maps, in addition to relationship and class annotations. Surgical datasets are inherently difficult to collect due to privacy regulations, restricted access to operating rooms, and strict hardware restrictions that must not interfere with procedures \cite{ozsoy2022_4D_OR, ozsoy2024mmor}. As a result, all existing external OR datasets rely on simulated procedures, and the specialized hardware they require (synchronized RGB-D sensors, calibration rigs) is unlikely to be tolerated in real surgical environments. Rather than addressing this gap, the field has moved in the opposite direction: MM-OR \cite{ozsoy2024mmor} introduces additional modalities for the same simulated procedure in the same OR, further increasing hardware requirements rather than reducing them to allow diverse real-world data collection. The annotation burden compounds this: constructing a new dataset for a different surgical procedure requires not only new recordings, but the full suite of multi-modal annotations. This limits real-world utility, scalability, and generalization.

Second, current methods cannot adapt to new clinical settings without substantial re-annotation. A knee replacement performed with robotic assistance involves different tools, configurations, and staff roles than a manual procedure \cite{henriques2025decoding}. Although it is the same procedure, existing methods would require an entirely new annotated dataset to handle the change, providing little incentive for adaptable methods. There is no mechanism to introduce new entity types such as a robotic arm, a specific instrument, or an additional staff role, without defining new annotations, retraining detection pipelines, and rebuilding the dataset. This rigidity creates a barrier to the development of new surgical technologies and methods that require rapid scientific iteration. We believe that this is the principal cause for the lack of diversity in external OR datasets for scene graphs that are currently limited to total-knee replacements performed in a single setting under identical conditions.   

To address these challenges, we propose SAGE-OR, a feature-centric surgical scene graph generation framework that replaces the detection-driven paradigm, where localization is a supervised prerequisite for reasoning, with a decoupled representation-reasoning paradigm in which localization is derived from frozen foundation models and encoded implicitly in pre-computed features. In contrast to existing approaches, SAGE-OR requires only RGB images and ground-truth relationship and class annotations during training, while preserving zero-shot localization information for downstream tasks. 

The core insight underlying SAGE-OR is that if sufficiently rich instance-level features can be pre-computed using accessible foundation models, then explicit localization supervision becomes unnecessary: correspondence between predicted and ground-truth entities can be learned implicitly via bipartite matching, spatial structure is preserved in the geometric feature space, and visual identity is captured by self-supervised representations. Prior work treats localization as supervision; we derive it and encode it in features without any supervision. Leveraging recent advances in zero-shot multi-view 3D reconstruction \cite{wang2024dust3r, wang2025vggt, wang2026pi3}, open-vocabulary panoptic segmentation \cite{carion2025sam3}, and large-scale visual representation learning \cite{simeoni2025dinov3}, we construct multi-view instance-level feature representations that implicitly encode spatial and semantic information. The resulting feature cache remains static during training and inference. The downstream graph transformer, with only 15M parameters, trains in 1.4 hours on a single commodity GPU, enabling rapid experimentation with architectures, encoders, and entity configurations. This design yields several advantages. First, by removing the dependency on dense supervision signals, SAGE-OR significantly reduces dataset construction complexity and improves adaptability to new surgical procedures by providing an incentive to create simple and diverse datasets not currently available. Second, the semi-supervised learning formulation allows the model to leverage unlabeled context: new entity types can be added to the segmentation prompts without additional annotation effort, and the downstream model will leverage them as unsupervised contextual nodes. Third, the decoupled architecture enables feature reuse across tasks and rapid research iteration: swapping visual encoders requires only regenerating the cache, not retraining the full pipeline. Extensive experiments demonstrate that SAGE-OR achieves 76\% F1 on the 4D-OR benchmark without any localization supervision and with a 4$\times$ smaller trainable model, and reaches 86\% F1, within 4 points of the SOTA method, through unsupervised hand augmentation.

In summary, our contributions are as follows:

(1) Semi-supervised learning without localization supervision: We propose a flexible training scheme based on Hungarian matching that eliminates the need for ground truth bounding boxes, human poses, segmentation masks, depth, and point cloud annotations. We demonstrate that this formulation enables unsupervised context augmentation, where the addition of unannotated entity types (\textit{e.g.}, hands) improves F1 by 10 points without any additional annotation, demonstrating a practical mechanism that could adapt to new clinical settings without additional annotations.

(2) Decoupled feature caching framework: We introduce a feature caching strategy that separates visual encoding from graph reasoning using only RGB input, requiring no depth sensors, camera calibration, or specialized hardware. Because Pi3 \cite{wang2026pi3} supports arbitrary camera configurations, including monocular input, the pipeline does not place a constraint on the number of views, reducing the barrier to real-world dataset construction using existing OR recording infrastructure. 

(3) Comprehensive encoder analysis: We provide the first systematic evaluation of modern visual encoders for surgical SGG, offering practical insights into the relationship between encoder choice and downstream performance.

(4) Efficient graph reasoning via decoupled inference: The decoupled architecture reduces online inference to a 15M-parameter graph transformer operating at approximately 1 ms per scene frame which consists of 6 images ($\sim$1000 FPS on cached features), allowing rapid experimentation to develop new models once new datasets become available.

(5) To the best of our knowledge, SAGE-OR is the first adaptable OR-SGG framework requiring only RGB input and sparse scene graph annotations while preserving spatial grounding, enabling deployment with standard recording equipment already present in clinical environments.

\section{Related Works}
\label{sec:related}
Below we provide a brief overview of relevant work grouped according to their paradigm. 

\textbf{Scene Graphs}
Scene graphs model objects as nodes and their interactions as edges, providing a structured representation for high-level scene understanding \cite{johnson2015image, krishna2017visual}. In the surgical domain, this paradigm has evolved from recognizing instrument-tissue interactions in endoscopic views \cite{nwoye2020recognition, nwoye2022rendezvous} to holistic operating room modeling \cite{henriques2025decoding} through action triplets $\langle subject, predicate, object \rangle$ \cite{nwoye2020recognition, rodin2024action}. Our approach differs from DETR-style set prediction \cite{carion2020detr} by utilizing coarse semantic constraints from foundation models to prune the bipartite search space, ensuring physically plausible assignments. ConceptGraphs \cite{gu2024conceptgraphs} builds object-centric 3D maps for navigation and Open-Vocabulary Video SGG \cite{Wuetal2024_OVVSGG} labels unseen predicates in 2D video, whereas our framework uses open-vocabulary entities as unsupervised anchors that enhance supervised surgical relations through 3D geometric grounding.

\textbf{Operating Room Datasets.}
The development of surgical SGG is predicated on the availability of high-fidelity datasets. Internal datasets such as CholecTriplet2021 \cite{nwoye2023cholectriplet} and Endoscapes \cite{murali2023endoscapes} focus on tool-tissue interactions, but lack the broader environmental context of the operating room. External datasets such as 4D-OR \cite{ozsoy2022_4D_OR} and MM-OR \cite{ozsoy2024mmor} provide a comprehensive view of the surgical theater but impose both an annotation and hardware burden: synchronized multi-view RGB-D streams, 3D bounding boxes, human poses, scene point clouds and object point clouds. Scaling to new procedures or operating rooms is therefore prohibitively expensive. By contrast, our method operates under a lightweight semi-supervised regime, requiring only images and scene graph annotation text files. 

\textbf{Surgical Scene Graphs.}
4D-OR \cite{ozsoy2022_4D_OR, ozsoy2021_MSSG} first localizes entities using PointNet++ \cite{qi2017pointnetplus}, followed by a relational reasoning module where the weights for each stage are trained independently. LABRAD-OR \cite{ozsoy2023_LABRAD_OR} introduced memory scene graphs for bimodal temporal reasoning. More recent single-stage approaches such as S$^2$Former-OR \cite{pei2025sgg} and TriTemp-OR \cite{guo2024trimodal} process images and point clouds simultaneously through end-to-end transformers and update all gradients in a single pass. Despite architectural differences, all existing methods share two constraints, repeated per-frame multi-modal encoding and dense localization supervision, whereas SAGE-OR treats localization as an implicit, pre-computed feature.

\textbf{Reducing Annotation Cost in Scene Graph Generation.}
Prior work on reducing SGG annotation cost outside the surgical domain falls into two main categories. Weakly-supervised image SGG (WS-ImgSGG) substitutes captions or LLM-extracted triplets for relation annotations \cite{zhong2021sgnls, ye2021lsws, zhang2023vs3, kim2024llm4sgg, yao2021visualds}, while weakly-supervised video SGG (WS-VidSGG) \cite{chen2023pla, kim2025nlvsgg, 
lee2026pals} focuses on generating scene graphs for video sequences rather than image frames. Both lines are 2D only and orthogonal to ours, reducing relation supervision while depending on category-trained detectors to produce localization proposals, which do not cover surgical entity ontologies and do not support multi-view 3D fusion required for external OR perspectives. These methods evaluate on natural-image distributions covered by their pretrained detectors (\textit{e.g.}, Visual Genome). Surgical scenes do not share this overlap and demand fine-grained disambiguation between visually similar clinicians and instruments that natural-image benchmarks do not exhibit. In the surgical domain, Pix2SG \cite{ozsoy2025pix2sg} introduced location-free SGG by predicting scene graphs as auto-regressive sequences, but discards spatial information entirely, preventing downstream tasks that require spatial grounding, a limitation acknowledged by the authors. SAGE-OR uses neither captions nor category-trained detectors, enabling prompt-level adaptation to new entity types and preserving full 3D spatial grounding. 

\textbf{Multi-modal Foundation Model Approaches in Operating Rooms.}
A separate line of work explores large-scale vision-language and multi-modal foundation models for OR understanding, integrating visual encoders with language models for high-level reasoning and instruction-driven analysis \cite{ozsoy2024mmor, ozsoy2024specialized, ozsoy2024oracle}. These approaches target open-ended semantic reasoning and multi-modal alignment by increasing dataset complexity for the same procedure. 

\section{Methodology}

\subsection{Cache Generation}
\textbf{3D Scene Representation Cache}
Rather than relying on calibrated depth sensors or supervised 3D detection as in prior work \cite{ozsoy2022_4D_OR, ozsoy2023_LABRAD_OR}, we obtain dense 3D structure from RGB alone using Pi3 \cite{wang2026pi3}, a zero-shot multi-view reconstruction model known not only to be orders of magnitude faster but also more robust, accurate, and versatile than procedural alternatives in challenging scenarios. Additionally, Pi3 consistently produces denser, more detailed point clouds that enable accurate centroid extraction for any entity in the scene (see Appendix B for a qualitative comparison). Given $N$ camera views per frame, Pi3 produces dense point maps of shape $H \times W \times 3$ per view, along with per-pixel confidence scores and camera poses. Each pixel is thus associated with a 3D coordinate in a local camera frame, which we transform to a shared world coordinate system using the poses. This provides a dense, per-pixel 3D representation of the scene without requiring depth sensors, camera calibration, or structure-from-motion \cite{schonberger2016sfm}. Because Pi3 estimates relative poses from arbitrary camera configurations, including monocular setups \cite{wang2026pi3}, SAGE-OR does not impose hardware requirements beyond standard RGB cameras already present in many ORs for recording, training, and legal documentation. The point maps enable cross-view instance fusion: by lifting 2D segmentation masks into 3D, we can determine whether a mask in one camera and a mask in another correspond to the same physical entity. In addition, they supply geometric features that are stored in the cache and later used for positional encoding and pairwise spatial reasoning.

\begin{figure}[t]
  \centering
  \includegraphics[width=\textwidth]{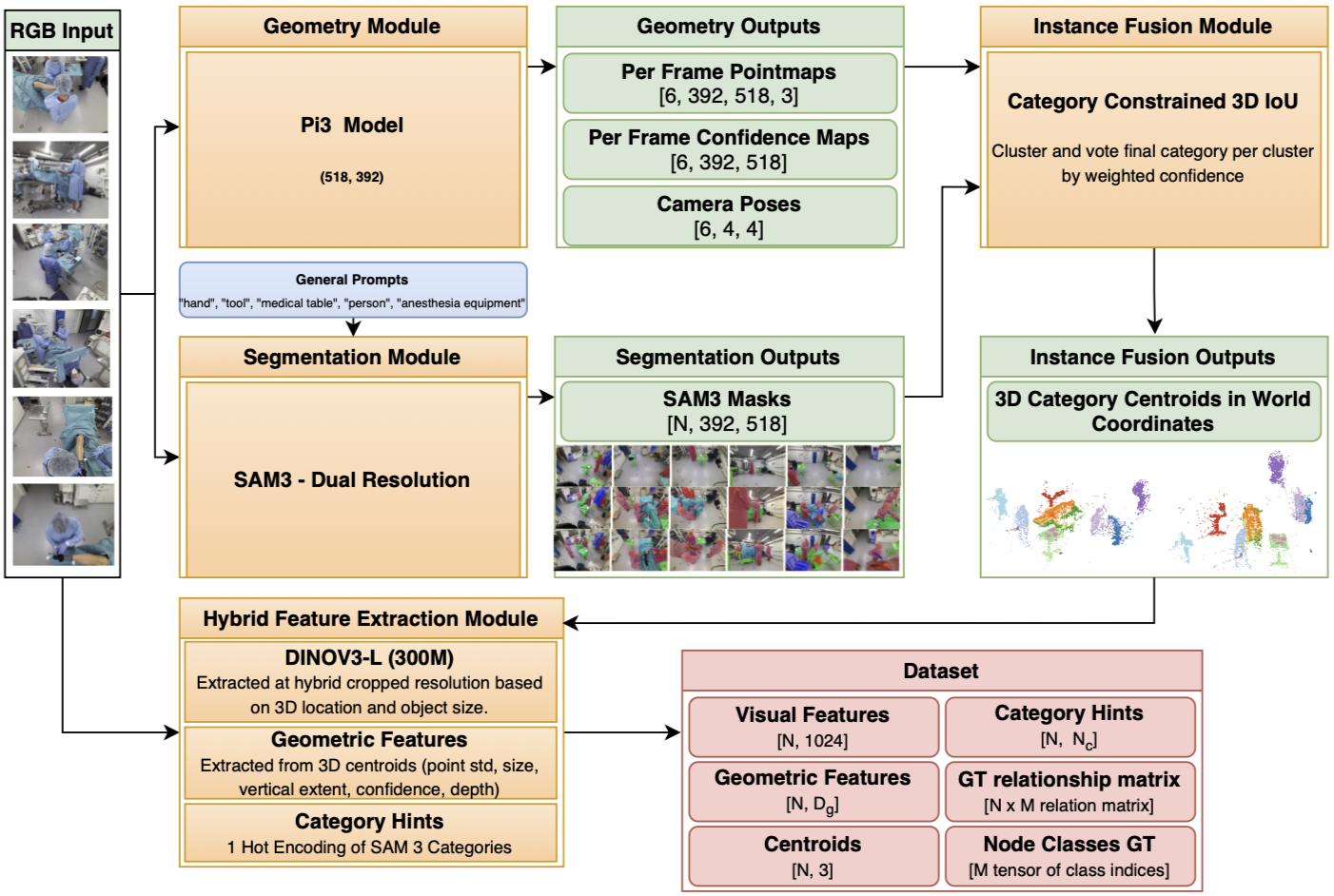}
  \caption{Overview of the 3-stage decoupled feature caching pipeline: 3D reconstruction (Pi3), open-vocabulary segmentation (SAM3), and instance fusion to produce a static per-frame cache of visual features, geometric features, centroids, and category hints. All stages use frozen models; the resulting cache is reused across all training and inference without modification. Bracketed values denote tensor shapes.}
  \label{fig:cache_pipeline}

\end{figure}

\textbf{Object Segmentation Cache}
We segment each camera view independently using SAM3 \cite{carion2025sam3}, an open-vocabulary panoptic segmentation model, prompted with general category names (\textit{e.g.}, \textit{person}, \textit{medical table}) rather than the fine-grained surgical classes present in the dataset. This choice is deliberate and empirically validated in exploratory experiments (see Appendices A.3 and A.4 for quantitative and qualitative analysis) with different prompt configurations using the training set class annotations to avoid test leakage. We found that general prompts consistently maximize segmentation recall. For example, \textit{tool} achieves  100\% recall while fine-grained alternatives degrade performance: \textit{surgical tool} achieves only 30\% recall, while \textit{saw}, \textit{scalpel}, and \textit{forceps} fail entirely at 0\%. In contrast, \textit{medical instrument} achieves 100\% recall but with severe over-detection, labeling nearly every medically related object in every frame. The same dual failure occurs for people: role-specific prompts like \textit{doctor}, \textit{nurse}, and \textit{surgeon} each label \textit{every} person in the scene including the patient, while \textit{assistant} achieves only 0.2\% recall. The general prompt \textit{person} captures all individuals without confusion and defers role disambiguation to the graph transformer (quantified in Appendix A.3). This recall-oriented strategy is central to our design: false positives (extra masks) simply produce unmatched nodes in Hungarian matching, where they either contribute useful context via attention or are effectively ignored. False negatives (missed entities) are the critical failure mode, as they cannot be recovered downstream. By optimizing upstream for recall and deferring precision to the learned graph transformer, the pipeline is robust to segmentation noise while remaining adaptable. Designing prompts for new procedures requires only identifying general categories, not fine-tuning prompt wording. Additional prompts (\textit{e.g.}, \textit{hand}, \textit{robotic arm}) can be introduced without any corresponding annotations. The downstream model leverages them as unsupervised contextual nodes through the semi-supervised formulation described in Section~\ref{sec:loss}. The process produces a set of binary masks per camera, each associated with a coarse category label. To resolve the many-to-many correspondence between masks across cameras, we perform category-constrained 3D instance fusion. For each mask, we extract the 3D points from the Pi3 point maps that fall within the mask region and cluster masks across cameras by computing 3D intersection-over-union (IoU) between their lifted point clouds, subject to the constraint that only masks of compatible categories may merge (see Appendix C for the full fusion algorithm, including category compatibility groups and adaptive scene scaling). This yields a set of fused instances per frame, each described by a coarse category, the set of cameras in which it is visible, a 3D centroid in world coordinates, a sampled point cloud, and an aggregate confidence score. These instances define the nodes of the scene graph. We use a confidence threshold of 0.1 for depth filtering, a 3D IoU threshold of 0.15 for cross-view merging, and a minimum density of 10 points to prune spurious detections. Because Pi3 reconstructs scenes up to an arbitrary global scale, we adaptively set the voxel size used for IoU computation to 2\% of the scene bounding-box diagonal, making the IoU threshold scale-invariant across procedures and camera configurations. These specific hyper-parameters were finalized through a grid search over confidence ranges ([0.0,0.5]), point density ([5,30]), and IoU ([0.05,0.3]) on the training partition, where we selected the configuration that maximized instance recall while maintaining graph stability. 

\textbf{Visual Feature Cache} 
Surgical instruments and hands typically occupy only a few pixels in the low-resolution view, which yields insufficient patch tokens for discrimination. To address this, we introduce a \textit{hybrid extraction strategy} using DINOv3 \cite{simeoni2025dinov3}. For large-entity categories (persons, tables, equipment), we extract features from the standard low-resolution input ($392 \times 518$) and pool within the mask, yielding typically 50-200 patches per instance. For small-entity categories (tools, hands), we crop the corresponding region from the high-resolution image ($1536 \times 2048$) with a 20\% bounding box padding, resize the crop to $518 \times 518$, and extract features from this enlarged view, producing approximately 300 patches per small object, a roughly 100$\times$ increase in feature density. For instances visible in multiple cameras, we aggregate across views via mean pooling after observing sub-optimal results from max-pooling. The complete per-frame cache comprises: visual features $\mathbf{v}_i \in \mathbb{R}^{d}$ (where $d$ depends on the encoder, \textit{e.g.}, 1024 for DINOv3-L), geometric features $\mathbf{g}_i \in \mathbb{R}^{D_g}$ derived from the instance's 3D point cloud, and category hints $\mathbf{c}_i \in \mathbb{R}^{N_c}$ for each of $N$ instances. At approximately 2 KB per instance in half-precision, an entire frame's cache is three orders of magnitude smaller than the raw multi-view images it replaces, making them suitable for edge hardware often used in the OR.

\subsection{Scene Graph Generation}

\begin{figure}[t]
  \centering
  \includegraphics[width=\textwidth]{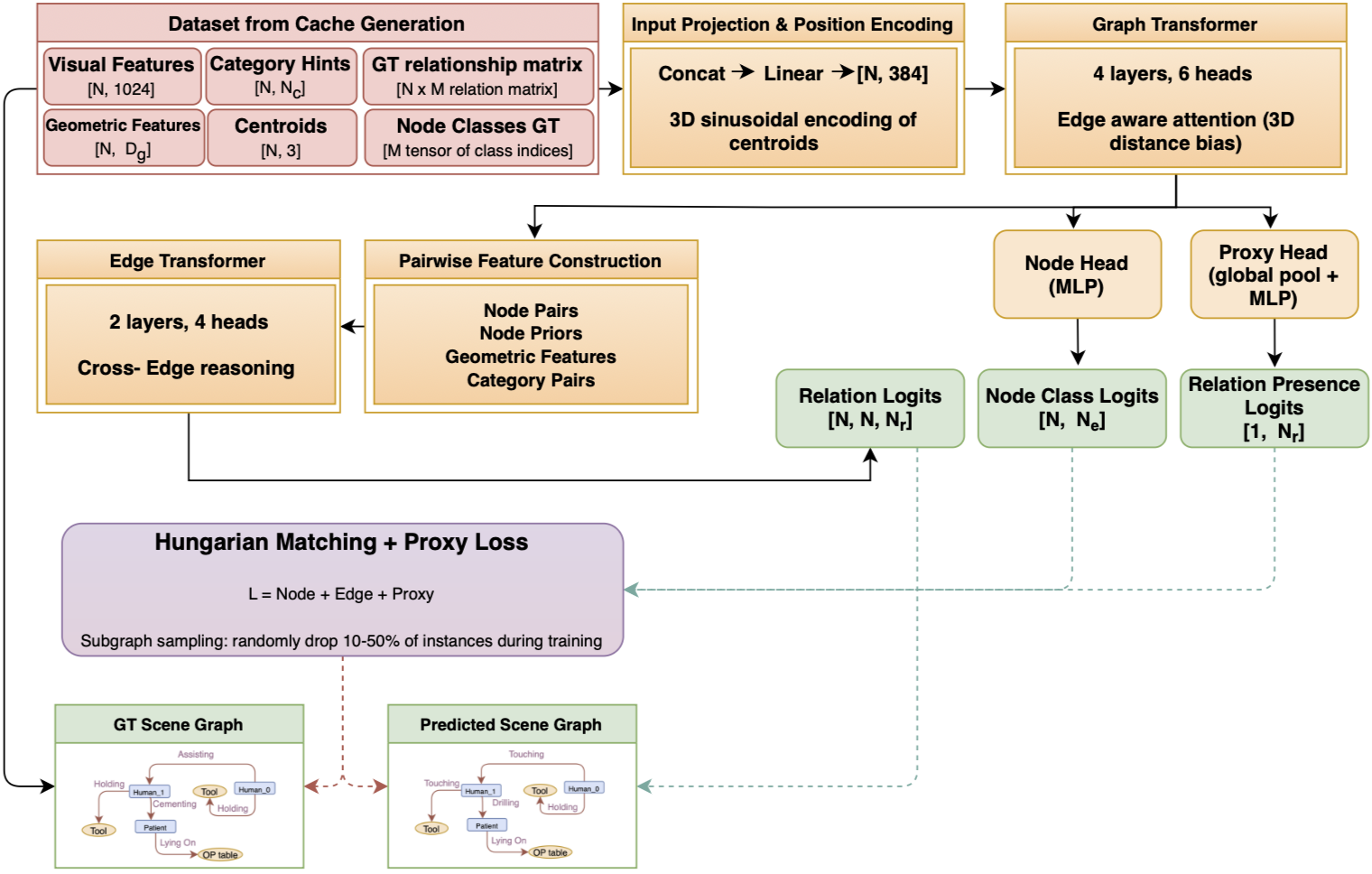}
  \caption{SAGE-OR training pipeline. Cached features are projected, positionally encoded, then processed by an edge-aware graph transformer. The node head produces entity classifications, while pairwise features are constructed and refined by an edge transformer for relation prediction. Hungarian matching aligns predicted nodes to ground-truth entities without localization supervision. Bracketed values denote tensor shapes.}
  \label{fig:training_pipeline}

  \vspace{-0.3cm}
\end{figure}

\textbf{Node Classification}
The node classification stage is where the deliberate gap between the cache's coarse categories and the task's fine-grained entity classes is resolved. The cache provides only broad labels \textit{person}, \textit{medical table}, \textit{tool}, while the scene graph requires specific identities such as \textit{human\_0} (head surgeon) or \textit{instrument\_table}. We intentionally defer this disambiguation to the learned model rather than encoding it in the segmentation prompts to circumvent any possible failure modes with frozen foundation models and increase adaptability, meaning the upstream cache requires no knowledge of the target ontology and can remain static across different procedures, operating rooms, or annotation schemes. Each cached instance is represented by the concatenation of its visual features, 3D centroid, and coarse category hint. A learned projection maps this combined representation to a shared embedding space, and we add a 3D sinusoidal positional encoding of the world-space centroid extended to three spatial dimensions (see Appendix E for the full formulation). The encoder then contextualizes each node through attention over all other nodes in the scene. The attention logits are augmented with a learned bias $\phi(d_{ij})$ derived from the pairwise Euclidean distance $d_{ij} = \|\mathbf{p}_i - \mathbf{p}_j\|_2$ between node centroids, so that nearby entities attend more strongly to each other, while distant pairs are down-weighted. This spatial inductive bias enables the model to disambiguate entities that are visually similar but occupy different roles because of their position. A lightweight 2-layer MLP maps each contextualized node embedding to class logits over the target entity categories. The predicted class distributions are also forwarded to the edge head with detached gradients as soft categorical priors. The node predictions are used for the edge classifier, but edge gradients do not flow back to the node classifier, providing the relation head with an explicit signal about what each node is likely to be without influencing the node classification, thereby keeping a clean separation of objectives.

\textbf{Edge Prediction}
Predicting relations requires reasoning over multiple complementary signals: a relation like \textit{Holding} depends on what the entities are, where they are, and what is visually happening between them. We construct a rich pairwise representation for each ordered node pair that fuses three sources of information. First, the pair of node embeddings from the graph transformer, along with their difference vector, which captures the relational contrast learned during node-level reasoning. Second, the set of soft node class priors produced by the node head (forwarded with detached gradients), which provides an explicit categorical signal. Third, pairwise geometry, computed from the cached geometric features. Formally, for a node pair $(i, j)$ with centroids $\mathbf{p}_i, \mathbf{p}_j$, bounding box dimensions $\mathbf{s}_i, \mathbf{s}_j$, and volumes $V_i, V_j$, the pairwise geometric descriptor $\mathbf{g}_{ij} \in \mathbb{R}^{D_{\text{p}}}$ is:
\begin{equation}
     \mathbf{g}_{ij} = \left[ d_{ij},\; \hat{\mathbf{r}}_{ij},\; \Delta z_{ij},\;
  \log\left(\frac{\mathbf{s}_j}{\mathbf{s}_i}+1\right),\;
  \log\left(\frac{V_j}{V_i}+1\right),\; \text{IoU}_{3D}(i,j),\;
  \frac{d_{ij}}{\bar{s}_{ij}} \right]
\end{equation}
where $d_{ij} = \|\mathbf{p}_i - \mathbf{p}_j\|_2$ is the Euclidean distance, $\hat{\mathbf{r}}_{ij}$ is the unit direction vector, $\Delta z_{ij}$ is the vertical height difference, $\bar{s}_{ij} = (\|\mathbf{s}_i\|+\|\mathbf{s}_j\|)/2$ is the mean box scale, and the remaining terms capture relative scale and overlap. Each component is independently normalized to zero mean and unit variance before concatenation. Rather than classifying each edge independently, we process the full set of pairwise features through a dedicated edge transformer, in which each edge attends to every other edge in the scene. This enables global consistency reasoning: if one clinician-patient pair is strongly predicted as \textit{Operating}, a second clinician holding an instrument can be more confidently classified as \textit{Assisting}. The edge transformer outputs relation logits over $N_r$ relation classes for each node pair.

\subsection{Semi-supervised Loss}
\label{sec:loss}

Since SAGE-OR eliminates localization supervision, cached instances do not have a known correspondence to ground-truth entities. We learn this correspondence end-to-end through Hungarian matching \cite{kuhn1955hungarian, carion2020detr}. Given $N$ predicted nodes and $M$ ground-truth entities, we construct a cost matrix $C \in \mathbb{R}^{N \times M}$ where the entry $C_{i,j}$ represents the matching cost between prediction $i$ and ground-truth $j$:
\begin{equation}
    C_{i,j} = \begin{cases} 1 - P(y_j | \mathbf{x}_i) & \text{if cat}(i, j) \text{ compatible} \\ \infty & \text{otherwise} \end{cases}
\end{equation}
where $P(y_j | \mathbf{x}_i)$ is the predicted probability for the ground-truth class and $\text{cat}(i,j)$ denotes the compatibility between the coarse SAM3 category of prediction $i$ and the ground-truth class of entity $j$ (\textit{e.g.}, a \textit{person} instance can match any human role but never an instrument). Losses and explicit geometric terms do not enter the cost, since ground-truth entities carry no localization. However, geometric and semantic information affect matching indirectly, as $P(y_j | \mathbf{x}_i)$ is computed from fused visual and geometric features under distance-biased attention, and the category constraint is a semantic prior. The Hungarian algorithm finds the optimal bipartite assignment $\mathcal{M}$ that minimizes $\sum_{(i,j) \in \mathcal{M}} C_{i,j}$. The total training objective combines three terms:
\begin{equation}
    \mathcal{L} = \lambda_{\text{node}} \mathcal{L}_{\text{node}} + \lambda_{\text{edge}} \mathcal{L}_{\text{edge}} + \lambda_{\text{proxy}} \mathcal{L}_{\text{proxy}},
\end{equation}
where $\mathcal{L}_{\text{node}}$ and $\mathcal{L}_{\text{edge}}$ are standard cross-entropy losses computed over the matched pairs in $\mathcal{M}$. The edge loss is weighted by the inverse class frequency to handle the high class imbalance in surgical datasets. To stabilize training for rare relations, we follow \cite{ozsoy2023_LABRAD_OR} and introduce a proxy loss $\mathcal{L}_{\text{proxy}}$, the efficacy of which is quantified in Table~\ref{tab:feature_ablation}, formulated as a frame-level multi-label binary cross-entropy task:
\begin{equation}
    \mathcal{L}_{\text{proxy}} = -\frac{1}{N_r} \sum_{r=1}^{N_r} [y_r \log(\sigma(\hat{y}_r)) + (1-y_r) \log(1-\sigma(\hat{y}_r))],
\end{equation}
with $\lambda_{\text{node}} = 1.0$, $\lambda_{\text{edge}} = 2.0$, and $\lambda_{\text{proxy}} = 1.0$ in all experiments (full formulation in Appendix E.6). Unmatched predictions receive no supervision, while unmatched ground-truth entities are treated as missed detections. Instances that remain unmatched still participate in the transformer's attention computation, providing contextual information to matched nodes without receiving direct supervision. This creates a node-level semi-supervised regime, distinct from classical semi-supervised learning over unlabeled samples: matched nodes receive supervision, while unmatched nodes contribute only as unlabeled context through attention. This mechanism enables a promising form of adaptability: new entity types can be added to the segmentation prompts without additional annotations. We show this in Section~\ref{sec:unsupervised_context}, where the addition of unannotated hand detections improves F1 by 10 points.

\section{Experiments}

\textbf{Experimental Setup and Dataset}
We evaluate SAGE-OR on the 4D-OR benchmark \cite{ozsoy2022_4D_OR}, the standard public dataset for OR SGG. We plan to extend our work to other holistic OR datasets as they become available. The dataset comprises 10 knee replacement surgery recordings totaling 6,734 frames with 6 images per frame, each captured from six synchronized RGB-D camera viewpoints at $1536 \times 2048$ resolution. Annotations include $N_e = 12$ entity classes and $N_r = 15$ relation classes (14 relations and none). The per-instance geometric features have dimension $D_{\text{g}} = 13$, pairwise geometric descriptors $D_{\text{p}} = 11$, and coarse category hints $N_c = 5$ representing the 5 segmentation prompts. We adopt the same data partitioning as prior work \cite{ozsoy2022_4D_OR, ozsoy2023_LABRAD_OR, pei2025sgg}: 6 takes (procedures) for training, 2 for validation, and 2 for testing. All results are reported using macro-averaged precision, recall, and F1 over the 15 relation classes on the held-out test set using the same evaluation protocol as prior work.

\textbf{Implementation Details}
\label{sec:implemenet}
SAGE-OR deliberately shifts computation from repeated per-frame encoding to a one-time offline pre-processing stage which competing methods cannot do because they require dense localization supervision and corresponding weight updates for all the encoders. While cache generation is non-trivial, the cost is parallelizable across views and frames, amortized across all downstream training runs and experiments, and reusable across different architectures and tasks. Caching is an optimization: features can be computed per frame at deployment (4.27s, dominated by SAM3; Table~\ref{tab:cache_timing}), so a multi-hour procedure accumulates no cache. This mirrors a broader trend where rich representations are pre-computed once and consumed by lightweight downstream models \cite{simeoni2025dinov3, carion2025sam3}.

\textbf{Model and Training.} The graph transformer uses a hidden dimension of 384, 4 encoder layers with 6 attention heads, and dropout of 0.3. The edge transformer uses 2 layers with 4 heads and a hidden dimension of 512 (see Appendix E for full architectural specifications, including edge-aware attention, node and edge classifiers, and the complete loss formulation). We train for up to 200 epochs for our final model with a batch size of 4 using AdamW with a learning rate of $3 \times 10^{-5}$, weight decay of $10^{-4}$, and a cosine schedule with 10 warmup epochs. All ablations are trained to 100 epochs for fair comparison. Subgraph sampling, which randomly drops detected instances during training to simulate occlusions and mitigate overfitting, uses a uniformly sampled keep ratio from $[0.5, 0.9]$. All experiments run on a single NVIDIA RTX Ada 6000 (48GB), but peak memory is under 2GB. Cache generation peaks at 6.4GB (Table~\ref{tab:cache_timing}) and remains a deployment consideration. The final 200 epoch model has approximately 15M trainable parameters and trains in approximately 1.4 hours.

\subsection{Comparison with State-of-the-Art}

Table~\ref{tab:sota} presents a per-class comparison against existing methods on the 4D-OR test set. All competing methods require 3D bounding boxes, 6D human poses, calibrated RGB-D sensors, object point clouds, and scene point clouds; S$^2$Former-OR also uses wrist coordinates from pose annotations to locate instruments \cite{pei2025sgg}. SAGE-OR requires only RGB images, entity class labels, and relationship annotations. Under this substantially reduced supervision regime, the core model achieves 76\% F1, matching the fully supervised 4D-OR baseline (75\%), and unsupervised hand augmentation (Section~\ref{sec:unsupervised_context}) raises this to 86\%, within 4 points of the best-performing, spatially grounded method. The trainable model is 4$\times$ smaller, trains in 1.4 hours, and processes cached features at approximately 1ms per frame. The 4-point gap decomposes cleanly: SAGE-OR (86\%) trails the fully supervised, non-temporal S$^2$Former-OR (89\%) by 3 points, and TriTemp-OR's temporal modeling contributes the remaining point, indicating that reduced supervision costs ${\approx}$3 F1 while the rest stems from the absence of temporal modeling. Detection recall adds no hidden cost: the final prompts achieve 100\% test recall for persons, anesthesia equipment, and tools, and 97\% for tables (Appendix~A.2). Assisting and Touching remain challenging across all methods regardless of supervision: fully supervised 4D-OR and S$^2$Former-OR score below our 62\% on Touching, indicating an inherent difficulty of the benchmark rather than a supervision artifact.

\begin{table*}[t]
\centering
\caption{Comparison with state-of-the-art OR-SGG methods on the 4D-OR test set. All values are percentages. Best F1 per class in \textbf{bold} among spatially grounded methods. Time/s reports seconds per scene following the convention of \cite{pei2025sgg}. $\ddagger$Temporal methods. Pix2SG (gray) is non-spatially grounded and autoregressive, reports average F1 only, and is listed for reference. $\dagger$Graph inference only; feature extraction is offline (4.27s/frame, Table~\ref{tab:cache_timing}).}
\label{tab:sota}
\setlength{\tabcolsep}{3pt}
\resizebox{\textwidth}{!}{%
\begin{tabular}{@{}l ccc l cccccccccccccc c@{}}
\toprule
Method & Params & Time/s & Supervision & & Assist & Cement & Clean & CloseTo & Cut & Drill & Hammer & Hold & LyingOn & Operate & Prepare & Saw & Suture & Touch & Avg \\
\midrule
\multirow{3}{*}{4D-OR \cite{ozsoy2022_4D_OR}} & \multirow{3}{*}{84.8M} & \multirow{3}{*}{1.28} & \multirow{3}{*}{Full}
 & P & 42 & 78 & 53 & 97 & 49 & 87 & 71 & 55 & 100 & 55 & 62 & 69 & 60 & 41 & 68 \\
 & & & & R & 93 & 78 & 63 & 89 & 49 & 100 & 89 & 95 & 99 & 99 & 91 & 91 & 100 & 69 & 87 \\ \rowcolor{gray!30}
 & & & & F1 & 58 & 78 & 57 & 93 & 49 & 93 & 79 & 70 & 99 & 71 & 74 & 79 & 75 & 51 & 75 \\
\midrule
\multirow{3}{*}{LABRAD-OR$^{\ddagger}$ \cite{ozsoy2023_LABRAD_OR}} & \multirow{3}{*}{85.8M} & \multirow{3}{*}{1.46} & \multirow{3}{*}{Full}
 & P & 60 & 96 & 86 & 96 & 91 & 100 & 93 & 71 & 100 & 85 & 77 & 78 & 100 & 71 & 87 \\
 & & & & R & 86 & 93 & 72 & 94 & 68 & 94 & 95 & 95 & 100 & 99 & 91 & 93 & 100 & 72 & 90 \\ \rowcolor{gray!30}
 & & & & F1 & 71 & 94 & 78 & \textbf{95} & 78 & \textbf{97} & 94 & 81 & \textbf{100} & 91 & 84 & 85 & \textbf{100} & 71 & 88 \\
\midrule
\multirow{3}{*}{S$^2$Former-OR \cite{pei2025sgg}} & \multirow{3}{*}{60.6M} & \multirow{3}{*}{0.98} & \multirow{3}{*}{Full}
 & P & 65 & 100 & 91 & 97 & 82 & 98 & 94 & 79 & 100 & 90 & 83 & 93 & 94 & 81 & 90 \\
 & & & & R & 66 & 94 & 86 & 93 & 85 & 92 & 94 & 83 & 100 & 91 & 89 & 99 & 97 & 49 & 88 \\ \rowcolor{gray!30}
 & & & & F1 & 66 & \textbf{97} & \textbf{89} & \textbf{95} & 84 & 95 & 94 & 81 & \textbf{100} & 91 & 86 & 96 & 95 & 61 & 89 \\
\midrule
\multirow{3}{*}{TriTemp-OR$^{\ddagger}$ \cite{guo2024trimodal}} & \multirow{3}{*}{67.1M} & \multirow{3}{*}{--} & \multirow{3}{*}{Full}
 & P & 74 & 100 & 92 & 97 & 86 & 98 & 96 & 84 & 100 & 93 & 86 & 95 & 96 & 82 & 91 \\
 & & & & R & 79 & 95 & 88 & 94 & 85 & 94 & 95 & 82 & 100 & 90 & 88 & 99 & 95 & 72 & 90 \\ \rowcolor{gray!30}
 & & & & F1 & \textbf{76} & \textbf{97} & \textbf{89} & \textbf{95} & \textbf{85} & 96 & \textbf{95} & \textbf{83} & \textbf{100} & \textbf{92} & \textbf{87} & \textbf{97} & 95 & \textbf{77} & \textbf{90} \\
\midrule
\textcolor{gray}{Pix2SG \cite{ozsoy2025pix2sg}} & \textcolor{gray}{--} & \textcolor{gray}{--} & \textcolor{gray}{Non-spatially grounded} & \textcolor{gray}{F1} & \textcolor{gray}{--} & \textcolor{gray}{--} & \textcolor{gray}{--} & \textcolor{gray}{--} & \textcolor{gray}{--} & \textcolor{gray}{--} & \textcolor{gray}{--} & \textcolor{gray}{--} & \textcolor{gray}{--} & \textcolor{gray}{--} & \textcolor{gray}{--} & \textcolor{gray}{--} & \textcolor{gray}{--} & \textcolor{gray}{--} & \textcolor{gray}{91} \\
\midrule \rowcolor{gray!30}
{SAGE-OR (core)} & 15M & $0.001^{\dagger}$ & Relations + classes & F1 & 54 & 71 & 77 & 92 & 63 & 83 & 79 & 79 & 100 & 87 & 77 & 62 & 69 & 56 & 76 \\
\midrule
\multirow{3}{*}{SAGE-OR (full, +hands)} & \multirow{3}{*}{15M} & \multirow{3}{*} {$0.001^{\dagger}$} & \multirow{3}{*}{Relations + classes}
 & P & 50 & 96 & 91 & 93 & 84 & 94 & 89 & 77 & 100 & 85 & 81 & 87 & 93 & 61 & 85 \\
 & & & & R & 61 & 96 & 80 & 92 & 75 & 94 & 93 & 87 & 100 & 92 & 79 & 87 & 96 & 63 & 86 \\ \rowcolor{gray!30}
 & & & & F1 & 55 & 96 & 85 & 93 & 80 & 94 & 91 & 82 & \textbf{100} & 88 & 80 & 87 & 94 & 62 & 86 \\
\bottomrule
\end{tabular}%
}
\vspace{-15pt}
\end{table*}

\subsection{Node Classification Performance}
\label{sec:node_perf}

We evaluate the efficacy of the coarse-to-fine disambiguation design by analyzing node classification independently. To evaluate node classification without localization data, we align predicted instances to ground-truth entities via bipartite Hungarian matching based on predicted class logits and SAM3 category constraints, followed by the standard classification report used in the 4D-OR benchmark. While the offline cache utilizes general-purpose prompts (\textit{e.g.}, \textit{person}, \textit{table}, \textit{tool}), our graph transformer must disambiguate these into the fine-grained surgical classes.  As shown in Table~\ref{tab:node_metrics}, the proposed graph transformer achieves 0.99 total node accuracy, with near-perfect F1-scores for primary entities. These results demonstrate that our architecture successfully leverages the rich representations extracted from the caching pipeline to resolve specific surgical roles and equipment types from generic category hints. This validates our decoupled design: because node identity is effectively solved by the graph transformer, the primary system bottleneck is isolated to relational reasoning rather than perception or entity disambiguation.

\subsection{Unsupervised Context Augmentation}
\label{sec:unsupervised_context}

A central claim of SAGE-OR is that the semi-supervised formulation allows unsupervised instances to improve predictions through contextual attention. Many surgical actions are physically performed with the hands. A clinician \textit{Cementing}, \textit{Suturing}, or \textit{Cutting} necessarily involves hand-tool-patient contact, yet the 4D-OR dataset contains no hand annotations. We hypothesize that, even without labels, hand detections provide a spatially and visually disambiguating context for action-oriented relations. We validate this directly by adding \textit{hand} as an additional SAM3 prompt category. The 4D-OR dataset contains no annotations for hands: no entity class, no relationship labels, no point clouds and no localization data. Hand instances are never matched by Hungarian matching and receive zero supervision during training, existing in the graph purely as contextual nodes.

Table~\ref{tab:hand_ablation} shows the per-class impact. Adding unsupervised hand nodes improves average F1 from 76\% to 86\%, a 10-point gain. The effect is particularly pronounced for action-oriented relations: \textit{Cementing} (+25), \textit{Suturing} (+25), \textit{Sawing} (+25), \textit{Cutting} (+17), and \textit{Hammering} (+12). These are precisely the relations where hand presence provides disambiguating context. Recall improves dramatically overall (75\% $\to$ 86\%), indicating that hand context helps the model detect relationships it would otherwise miss entirely.

\begin{table*}[t]
\centering
\caption{Per-class impact of adding unsupervised hand nodes on the 4D-OR test set. Despite receiving no supervision, hand nodes improve average F1 by +10 points. $\Delta$ denotes absolute F1 change per class.}
\label{tab:hand_ablation}
\setlength{\tabcolsep}{3pt}
\resizebox{\textwidth}{!}{%
\begin{tabular}{@{}l l cccccccccccccc c@{}}
\toprule
Configuration & & Assist & Cement & Clean & CloseTo & Cut & Drill & Hammer & Hold & LyingOn & Operate & Prepare & Saw & Suture & Touch & Avg \\
\midrule
\multirow{3}{*}{Without hands} & P & 44 & 100 & 84 & 92 & 93 & 96 & 70 & 77 & 99 & 83 & 78 & 79 & 74 & 54 & 81 \\
 & R & 68 & 56 & 71 & 92 & 47 & 73 & 91 & 82 & 100 & 92 & 77 & 51 & 65 & 57 & 75 \\ \rowcolor{gray!30}
 & F1 & 54 & 71 & 77 & 92 & 63 & 83 & 79 & 79 & \textbf{100} & 87 & 77 & 62 & 69 & 56 & 76 \\
\midrule
\multirow{3}{*}{With hands (final)} & P & 50 & 96 & 91 & 93 & 84 & 94 & 89 & 77 & 100 & 85 & 81 & 87 & 93 & 61 & 85 \\
 & R & 61 & 96 & 80 & 92 & 75 & 94 & 93 & 87 & 100 & 92 & 79 & 87 & 96 & 63 & 86 \\
 \rowcolor{gray!30}
& F1 & \textbf{55} & \textbf{96} & \textbf{85} & \textbf{93} & \textbf{80} & \textbf{94} & \textbf{91} & \textbf{82} & \textbf{100} & \textbf{88} & \textbf{80} & \textbf{87} & \textbf{94} & \textbf{62} & \textbf{86} \\

\midrule
 & $\Delta$ F1 & +1 & +25 & +8 & +1 & +17 & +11 & +12 & +3 & 0 & +1 & +3 & +25 & +25 & +6 & +10 \\
\bottomrule
\end{tabular}%
}
\vspace{-15pt}
\end{table*}

This result validates the semi-supervised mechanism and provides a concrete demonstration of the adaptability that the decoupled caching design enables. The adaptation workflow did not require additional annotation effort. We added a single word to the SAM3 prompt list, regenerated the cache, and retrained without changes to the model architecture or loss function.
More broadly, this experiment illustrates within-benchmark adaptation, introducing unannotated entities in a known setting. Adapting to a genuinely new procedure with different entity ontologies and relation types is a distinct requirement that we leave to future work. A practitioner can hypothesize that a new entity type (a robotic arm, a specific instrument) might provide useful context, add it to the prompt list, and evaluate the impact, all without modifying annotations or retraining the segmentation model. The static caching design makes this iteration loop practical: regenerating the cache for a new prompt takes a few hours, and the downstream model trains in 1.4 hours. This capability is architecturally unavailable for existing methods. In LABRAD-OR, S$^2$Former-OR, and TriTemp-OR, every entity in the scene graph must pass through a fully supervised detection pipeline: adding hands would require full multi-modal hand annotations in every frame and retraining the detection backbone. There is no mechanism for an entity to participate in graph reasoning without ground-truth labels flowing through every stage of the pipeline. SAGE-OR is the only method in which an entity can influence relational predictions with zero supervision, because Hungarian matching simply leaves it unmatched while attention propagates its information to all other nodes. These entities refine predictions over the annotated ontology rather than introducing new output classes.

To verify that the model selectively leverages informative context rather than benefiting from additional nodes indiscriminately, we conduct a control study with three prompts that are irrelevant to the task and existing graph nodes: \textit{knee} (already captured as part of the patient entity), \textit{door} (OR infrastructure unrelated to surgical actions), and \textit{shoe} (detected on clinicians' feet). All prompts were validated to produce sufficient instance detections as shown in Table~\ref{tab:control}. More importantly, Table~\ref{tab:control} shows that none of the control prompts improves or degrades performance, confirming that the +10 F1 gain from hands reflects a genuinely disambiguating signal rather than an artifact of increased graph density. To address the scope of our contextual augmentation, it is critical to note that \textit{hands} are the only viable candidates for this experiment. After an exhaustive review of the scene dynamics, we found no other unannotated entities that contribute to action disambiguation. Seemingly relevant prompts, such as \textit{knee}, are strictly redundant; the patient is already localized with 100\% recall, rendering sub-components uninformative for the downstream graph transformer.

\begin{table}[t]
\begin{minipage}[t]{0.50\textwidth}
\centering
\caption{Fine-grained node classification metrics. Near-perfect performance on primary entities validates that frozen features are not a bottleneck for identity disambiguation.}
\label{tab:node_metrics}
\scriptsize
\begin{tabular}{@{}l ccc r@{}}
\toprule
Node Entity & P & R & F1 & Support \\
\midrule
Patient & 0.99 & 1.00 & 0.99 & 1210 \\
Anesthesia Equip. & 1.00 & 1.00 & 1.00 & 1355 \\
Human\_0 & 0.99 & 1.00 & 0.99 & 1263 \\
Human\_1 & 1.00 & 1.00 & 1.00 & 1108 \\
Human\_2 & 0.97 & 0.99 & 0.98 & 788 \\
Human\_3 & 0.94 & 0.64 & 0.76 & 101 \\
Human\_4 & 0.00 & 0.00 & 0.00 & 2 \\
Human\_5 & 0.00 & 0.00 & 0.00 & 0 \\
Instrument & 1.00 & 1.00 & 1.00 & 1377 \\
Instrument Table & 1.00 & 1.00 & 1.00 & 1367 \\
Operating Table & 1.00 & 1.00 & 1.00 & 1155 \\
Secondary Table & 1.00 & 1.00 & 1.00 & 1377 \\
\midrule
\textbf{Total Accuracy} & \multicolumn{3}{c}{--} & \textbf{0.99} \\

\bottomrule
\end{tabular}
\end{minipage}
\hfill
\begin{minipage}[t]{0.47\textwidth}
\centering

\caption{Prompt selectivity control study. Irrelevant prompts do not improve or degrade over the
hands-only baseline.}
\label{tab:control}
\scriptsize
\begin{tabular}{@{}l c c c@{}}
\toprule
Configuration & Instances & F1 (\%) & $\Delta$ \\
\midrule
Hands only (base) & 31,673 & 82 & -- \\
+ knee (irrelevant) & 5,916 & 82 & 0 \\
+ door (irrelevant) & 29,905 & 82 & 0 \\
+ shoe (irrelevant) & 32,264 & 82 & 0 \\
\bottomrule
\end{tabular}

\vspace{0.6cm} 

\caption{Amortized one-time cache generation cost on a single RTX Ada 6000.}
\label{tab:cache_timing}
\scriptsize
\begin{tabular}{@{}l ccc c@{}}
\toprule
Stage & Per Frame & Total & Peak Mem & \% \\
\midrule
SAM3 & 2.58s & 4.92h & 5097 MB & 61\% \\
Pi3 & 0.56s & 1.05h & 6464 MB & 13\% \\
DINOv3 & 1.13s & 2.12h & 1736 MB & 26\% \\
\midrule
\textbf{Total} & \textbf{4.27s} & \textbf{8.1h} & \textbf{6464 MB} & 100\% \\
\bottomrule
\end{tabular}
\vspace{0.1cm}
\end{minipage}
\end{table}

\subsection{Encoder Analysis}

The decoupled caching design allows the visual encoder to be swapped by regenerating only the cache. We provide the first systematic comparison of modern visual encoders for surgical SGG, evaluating six encoders that span the self-supervised (DINOv3), contrastive (CLIP), and supervised (ResNet-50, EfficientNet) paradigms. All experiments use identical downstream hyper-parameters. Table~\ref{tab:encoders} presents a central finding: DINOv3 ViT-L achieves 82\% F1, 7--16 points above non-DINOv3 alternatives, suggesting that self-supervised features encode richer instance-level semantics than representations trained for classification or image-text alignment. DINOv3 ViT-7B drops to 67\% F1 despite producing 4$\times$ larger features, likely due to overfitting on the small training set. The best encoder is not the largest, but the one that best balances feature quality with downstream trainability. EfficientNet-B5-NS and CLIP ViT-L, the encoders used by LABRAD-OR \cite{ozsoy2023_LABRAD_OR} and MM-OR \cite{ozsoy2024mmor} respectively, achieve only 74--75\% F1 in our framework matching 4D-OR performance.

\begin{table}[t]
\begin{minipage}[t]{0.50\textwidth}
\centering
\caption{Encoder comparison on 4D-OR test set. All models use identical downstream configuration.}
\label{tab:encoders}
\scriptsize
\begin{tabular}{@{}l ccc c@{}}
\toprule
Encoder & P & R & F1 & Mem \\
\midrule
DINOv3 ViT-L/16 & 78 & 88 & \textbf{82} & 1784 \\
DINOv3 ViT-B/16 & 76 & 88 & 81 & 1768 \\
DINOv3 ViT-S/16 & 74 & 82 & 77 & 1730 \\
DINOv3 ViT-7B/16 & 72 & 67 & 67 & 3056 \\
CLIP ViT-L/14 & 74 & 78 & 75 & 1784 \\
EffNet-B5-NS & 74 & 76 & 74 & 1634 \\
ResNet-50 & 77 & 62 & 66 & 1964 \\
\bottomrule
\end{tabular}
\end{minipage}
\hfill
\begin{minipage}[t]{0.48\textwidth}
\centering
\caption{Feature ablation on 4D-OR test set. Each row adds a component to the base model.}
\label{tab:feature_ablation}
\scriptsize
\begin{tabular}{@{}l ccc@{}}
\toprule
Configuration & P & R & F1 \\
\midrule
Base (node+geom) & 76 & 87 & 82 \\
\quad + Cat. hints & 79 & 84 & 81 \\
\quad + Obj. types & 80 & 86 & 82 \\
\quad + Class priors & 78 & 86 & 81 \\
\quad + Pair. geom. & 76 & 88 & 81 \\
\quad + Types+priors & 79 & 86 & 82 \\
\quad + All fused & 76 & 85 & 80 \\
All features & 79 & 88 & 82 \\
All (35M params) & 80 & 87 & 82 \\
All (no proxy) & 79 & 84 & 80 \\
\bottomrule
\end{tabular}
\end{minipage}
\vspace{-15pt}
\end{table}

\subsection{Feature and Augmentation Ablations}
\label{sec:visual_ablation}

\begin{table}[t]
\begin{minipage}[t]{0.52\textwidth}
\centering
\caption{Visual feature ablation.}
\label{tab:visual_ablation}
\small
\begin{tabular}{@{}l ccc c@{}}
\toprule
Configuration & P & R & F1 & Mem \\
\midrule
All features & 77 & 89 & 82 & 1756 \\
No visual edge & 76 & 90 & 82 & 1554 \\
No visual node & 50 & 62 & 53 & 1784 \\
Geometry only & 50 & 63 & 54 & 1578 \\
\bottomrule
\end{tabular}
\end{minipage}
\hfill
\begin{minipage}[t]{0.45\textwidth}
\centering
\caption{Augmentation ablation.}
\label{tab:aug_ablation}
\small
\begin{tabular}{@{}l ccc@{}}
\toprule
Augmentation & P & R & F1 \\
\midrule
None & 78 & 83 & 80 \\
Subgraph sampling & 79 & 87 & 82 \\
+ node dropout & 75 & 87 & 80 \\
+ feature dropout & 74 & 85 & 78 \\
\bottomrule
\end{tabular}
\end{minipage}

\end{table}

Table~\ref{tab:feature_ablation} evaluates each feature component in the edge prediction head. Doubling the model to 35M parameters provides no F1 benefit, confirming the bottleneck is feature quality rather than capacity. Removing the proxy loss reduces F1 by 2 points, validating auxiliary supervision for rare relations. The All-features configuration consistently shows less overfitting, which allows for longer training and therefore, increased performance in the final model. Table~\ref{tab:visual_ablation} shows a non-obvious finding: visual features are redundant in the edge head, as their removal leaves F1 unchanged while saving 200 MB and 2M parameters and gaining 200 FPS, motivating their exclusion from the final model. Removing visual features from the node input causes a catastrophic drop to 53\%, confirming that DINOv3 features are essential for entity disambiguation, while geometric features alone cannot substitute for visual identity. In Table~\ref{tab:aug_ablation}, we show that subgraph sampling provides a 2-point F1 improvement driven by a 4-point recall gain, confirming its role as an effective regularizer. The subgraph sampling strategy additionally results in 5$\times$ less overfitting compared to the base model, allowing training for extended periods with continued performance gains in the final model.

\section{Conclusion}

We presented SAGE-OR, a scene graph generation framework for operating rooms that decouples visual feature extraction from graph reasoning through offline pre-computation using accessible frozen foundation models. By replacing dense multi-modal supervision with a semi-supervised formulation based on Hungarian matching and general-purpose prompts, SAGE-OR eliminates the need for bounding boxes, human poses, depth maps, and point cloud annotations while achieving 76\% F1 on the 4D-OR benchmark, rising to 86\% with unsupervised hand augmentation, with a 4$\times$ smaller trainable model than prior work. Our unsupervised context augmentation experiment demonstrates a practical mechanism for adaptation without additional annotations. Training completes in 1.4 hours and relational inference requires approximately 1ms per frame on pre-computed features, enabling the rapid iteration over architectures and entity configurations needed to develop new technologies.

\textbf{Discussion and Future Work.} Following all prior OR scene graph generation methods \cite{ozsoy2022_4D_OR, ozsoy2023_LABRAD_OR, pei2025sgg, guo2024trimodal}, we evaluate exclusively on the 4D-OR benchmark. The scarcity of external OR scene graph benchmarks is a direct consequence of two barriers that SAGE-OR is designed to address. The first is annotation burden. The second is hardware burden: existing datasets require synchronized RGB-D sensors and calibration infrastructure that will not be tolerated in real surgical environments, where any disruption to clinical staff is unacceptable. Pi3 generates 3D structure from arbitrary uncalibrated camera configurations \cite{wang2026pi3}, meaning datasets could be constructed from the simple recording equipment already present in many ORs. In future work, we aim to evaluate SAGE-OR on a wider variety of surgical procedures and OR layouts as new holistic datasets become available. MM-OR shows added modalities contribute ${\approx}$4 F1 in complex settings \cite{ozsoy2024mmor}, and our claim is not that they cannot help, but that competitive accuracy is attainable without their acquisition burden. To isolate the impact of our feature-centric spatial reasoning, we intentionally adopt a robust per-frame formulation. While temporal modeling is a natural extension for modeling long-range procedural dependencies, our results demonstrate that high-fidelity foundation features alone provide a sufficient signal for complex relational disambiguation. Moreover, to move past the current offline caching bottleneck, we plan to explore lightweight, on-the-fly feature extraction and segmentation to enable end-to-end real-time inference.

\section*{Acknowledgment}
\label{sec:ack}

This research was undertaken, in part, based on support from the Natural Sciences and Engineering Research Council of Canada Grant RGPIN-2021-03479 (NSERC DG) and through the Natural Sciences and Engineering Research Council of Canada CGS-M scholarship (NSERC CGS-M).

\clearpage

\bibliography{references}

@STRING{cvpr	= {Proceedings of the IEEE/CVF Conference on Computer Vision
		  and Pattern Recognition} }

@STRING{cvprw	= {Proceedings of the IEEE/CVF Conference on Computer Vision
		  and Pattern Recognition Workshops} }

@STRING{eccv	= {Proceedings of the European Conference on Computer
		  Vision} }

@STRING{iccv	= {Proceedings of the IEEE/CVF International Conference on
		  Computer Vision} }

@STRING{iclr	= {Proceedings of the International Conference on Learning
		  Representations} }

@STRING{icra	= {Proceedings of the IEEE International Conference on
		  Robotics and Automation} }

@STRING{ijcv	= {International Journal of Computer Vision} }

@STRING{neurips	= {Advances in Neural Information Processing Systems} }

@STRING{miccai	= {Proceedings of the International Conference on Medical Image Computing and Computer-Assisted Intervention} }

@Article{	  simeoni2025dinov3,
  author	= {Sim{\'e}oni, Oriane and Vo, Huy V. and Seitzer, Maximilian
		  and Baldassarre, Federico and Oquab, Maxime and Jose, Cijo
		  and Khalidov, Vasil and Szafraniec, Marc and Yi, Seungeun
		  and Ramamonjisoa, Micha{\"e}l and others},
  title		= {{DINOv3}},
  journal	= {arXiv preprint arXiv:2508.10104},
  year		= {2025}
}

@InProceedings{	  wang2025vggt,
  title		= {{VGGT}: Visual geometry grounded transformer},
  author	= {Wang, Jianyuan and Chen, Minghao and Karaev, Nikita and
		  Vedaldi, Andrea and Rupprecht, Christian and Novotny,
		  David},
  booktitle	= cvpr,
  pages		= {5294--5306},
  year		= {2025}
}

@InProceedings{	  wang2026pi3,
  author	= {Wang, Yifan and Zhou, Jianjun and Zhu, Haoyi and Chang,
		  Wenzheng and Zhou, Yang and Li, Zizun and Chen, Junyi and
		  Pang, Jiangmiao and Shen, Chunhua and He, Tong},
  title		= {{$\pi^3$}: Permutation-Equivariant Visual Geometry
		  Learning},
  booktitle	= iclr,
  year		= {2026}
}

@InProceedings{	  schonberger2016sfm,
  title		= {Structure-from-motion revisited},
  author	= {Sch{\"{o}}nberger, Johannes Lutz and Frahm, Jan-Michael},
  booktitle	= cvpr,
  year		= {2016},
  pages		= {4104--4113}
}

@InProceedings{	  wang2024dust3r,
  title		= {{DUSt3R}: Geometric {3D} vision made easy},
  author	= {Wang, Shuzhe and Leroy, Vincent and Cabon, Yohann and
		  Chidlovskii, Boris and Revaud, Jerome},
  booktitle	= cvpr,
  year		= {2024},
  pages		= {20697--20709}
}

@InProceedings{	  gu2024conceptgraphs,
  title		= {{ConceptGraphs}: Open-vocabulary {3D} scene graphs for
		  perception and planning},
  author	= {Gu, Qiao and Kuwajerwala, Ali and Morin, Sacha and
		  Jatavallabhula, Krishna Murthy and Sen, Bipasha and
		  Agarwal, Aditya and Rivera, Corban and Paul, William and
		  Ellis, Kirsty and Chellappa, Rama and others},
  booktitle	= icra,
  pages		= {5021--5028},
  year		= {2024}
}

@InProceedings{	  ozsoy2022_4d_or,
  author	= {{\"O}zsoy, Ege and {\"O}rnek, Evin P{\i}nar and Eck,
		  Ulrich and Czempiel, Tobias and Tombari, Federico and
		  Navab, Nassir},
  title		= {{4D-OR}: Semantic Scene Graphs for {OR} Domain Modeling},
  booktitle	= miccai,
  pages		= {475--485},
  organization	= {Springer},
  year		= {2022}
}

@article{pei2025sgg,
  title={{S$^2$Former-OR}: Single-Stage Bi-Modal Transformer for Scene Graph Generation in {OR}},
  author={Pei, Jialun and Guo, Diandian and Zhang, Jingyang and Lin, Manxi and Jin, Yueming and Heng, Pheng-Ann},
  journal={IEEE Transactions on Medical Imaging},
  volume={44},
  number={1},
  pages={361--372},
  year={2024},
  publisher={IEEE}
}

@inproceedings{ozsoy2023_labrad_or,
  title={{LABRAD-OR}: Lightweight memory scene graphs for accurate bimodal reasoning in dynamic operating rooms},
  author={{\"O}zsoy, Ege and Czempiel, Tobias and Holm, Felix and Pellegrini, Chantal and Navab, Nassir},
  booktitle=miccai,
  pages={302--311},
  year={2023},
  organization={Springer}
}

@Article{	  ozsoy2021_mssg,
  author	= {{\"O}zsoy, Ege and {\"O}rnek, Evin P{\i}nar and Eck,
		  Ulrich and Tombari, Federico and Navab, Nassir},
  title		= {Multimodal Semantic Scene Graphs for Holistic Modeling of
		  Surgical Procedures},
  journal	= {arXiv preprint arXiv:2106.15309},
  year		= {2021}
}

@inproceedings{ozsoy2024mmor,
  title={{MM-OR}: A large multimodal operating room dataset for semantic understanding of high-intensity surgical environments},
  author={{\"O}zsoy, Ege and Pellegrini, Chantal and Czempiel, Tobias and Tristram, Felix and Yuan, Kun and Bani-Harouni, David and Eck, Ulrich and Busam, Benjamin and Keicher, Matthias and Navab, Nassir},
  booktitle=cvpr,
  pages={19378--19389},
  year={2025},
  organization={IEEE}
}

@inproceedings{carion2025sam3,
  title={{SAM 3}: Segment anything with concepts},
  author={Carion, Nicolas and Gustafson, Laura and Hu, Yuan-Ting and Debnath, Shoubhik and Hu, Ronghang and Suris Coll-Vinent, Didac and Ryali, Chaitanya and Alwala, Kalyan Vasudev and Khedr, Haitham and Huang, Andrew and others},
  booktitle=iclr,
  volume={2026},
  pages={138846--138923},
  year={2026}
}

@InProceedings{	  qi2017pointnetplus,
  author	= {Qi, Charles R. and Yi, Li and Su, Hao and Guibas, Leonidas
		  J.},
  title		= {{PointNet++}: Deep Hierarchical Feature Learning on Point
		  Sets in a Metric Space},
  booktitle	= neurips,
  year		= {2017}
}

@article{   alhajj2019cataracts,
  title={{CATARACTS}: Challenge on automatic tool annotation for {cataRACT} surgery},
  author={Hassan {Al Hajj} and Mathieu Lamard and Pierre-Henri Conze and Soumali Roychowdhury and Xiaowei Hu and Gabija Mar{\v{s}}alkait{\.{e}} and Odysseas Zisimopoulos and Muneer Ahmad Dedmari and Fenqiang Zhao and Jonas Prellberg and Manish Sahu and Adrian Galdran and Teresa Ara{\'u}jo and Duc My Vo and Chandan Panda and Navdeep Dahiya and Satoshi Kondo and Zhengbing Bian and Arash Vahdat and Jonas Bialopetravi{\v{c}}ius and Evangello Flouty and Chenhui Qiu and Sabrina Dill and Anirban Mukhopadhyay and Pedro Costa and Guilherme Aresta and Senthil Ramamurthy and Sang-Woong Lee and Aur{\'e}lio Campilho and Stefan Zachow and Shunren Xia and Sailesh Conjeti and Danail Stoyanov and Jogundas Armaitis and Pheng-Ann Heng and William G. Macready and B{\'e}atrice Cochener and Gwenol{\'e} Quellec},
  journal={Medical Image Analysis},
  year={2019},
  volume={52},
  pages={24--41}
}

@article{allan2020robotic,
  title={2018 robotic scene segmentation challenge},
  author={Allan, Max and Kondo, Satoshi and Bodenstedt, Sebastian and Leger, Stefan and Kadkhodamohammadi, Rahim and Luengo, Imanol and Fuentes, Felix and Flouty, Evangello and Mohammed, Ahmed and Pedersen, Marius and others},
  journal={arXiv preprint arXiv:2001.11190},
  year={2020}
}

@Article{	  ban2024concept,
  author	= {Ban, Yutong and Eckhoff, J. A. and Ward, T. M. and
		  Hashimoto, D. A. and Meireles, O. R. and Rus, Daniela and
		  Rosman, Guy},
  title		= {Concept Graph Neural Networks for Surgical Video
		  Understanding},
  journal	= {IEEE Transactions on Medical Imaging},
  volume	= {43},
  number	= {1},
  pages		= {264--274},
  year		= {2024}
}

@Article{	  ding2025visual,
  author	= {Ding, Di and Yao, Tianliang and Luo, Rong and Sun, Xusen},
  title		= {Visual Question Answering in Robotic Surgery: A
		  Comprehensive Review},
  journal	= {IEEE Access},
  volume	= {13},
  pages		= {9473--9484},
  year		= {2025}
}

@InProceedings{	  guo2024trimodal,
  author	= {Guo, Diandian and Lin, Manxi and Pei, Jialun and Tang, He
		  and Jin, Yueming and Heng, Pheng-Ann},
  title		= {Tri-modal Confluence with Temporal Dynamics for Scene
		  Graph Generation in Operating Rooms},
  booktitle	= miccai,
  pages		= {714--724},
  year		= {2024}
}

@InProceedings{	  johnson2015image,
  author	= {Johnson, Justin and Krishna, Ranjay and Stark, Michael and
		  Li, Li-Jia and Shamma, David A. and Bernstein, Michael S.
		  and Fei-Fei, Li},
  title		= {Image Retrieval Using Scene Graphs},
  booktitle	= cvpr,
  pages		= {3668--3678},
  year		= {2015}
}

@Article{	  krishna2017visual,
  author	= {Krishna, Ranjay and Zhu, Yuke and Groth, Oliver and
		  Johnson, Justin and Hata, Kenji and Kravitz, Joshua and
		  Chen, Stephanie and Kalantidis, Yannis and Li, Li-Jia and
		  Shamma, David A. and Bernstein, Michael S. and Fei-Fei,
		  Li},
  title		= {{Visual Genome}: Connecting Language and Vision Using
		  Crowdsourced Dense Image Annotations},
  journal	= ijcv,
  volume	= {123},
  number	= {1},
  pages		= {32--73},
  year		= {2017}
}

@Article{	  moglia2021systematic,
  author	= {Moglia, A. and Georgiou, K. and Georgiou, E. and Satava,
		  R. M. and Cuschieri, A.},
  title		= {A Systematic Review on Artificial Intelligence in
		  Robot-assisted Surgery},
  journal	= {International Journal of Surgery},
  volume	= {95},
  pages		= {106151},
  year		= {2021}
}

@Article{	  murali2023endoscapes,
  author	= {Murali, Aditya and Alapatt, Deepak and Mascagni, Pietro
		  and Vardazaryan, Armine and Garcia, Alain and Okamoto,
		  Nariaki and Costamagna, Guido and Mutter, Didier and
		  Marescaux, Jacques and Dallemagne, Bernard and Padoy,
		  Nicolas},
  title		= {The {Endoscapes} dataset for surgical scene segmentation,
		  object detection, and critical view of safety assessment:
		  Official splits and benchmark},
  journal	= {arXiv preprint arXiv:2312.12429},
  year		= {2023}
}

@Article{	  murali2023latent,
  title		= {Latent graph representations for critical view of safety
		  assessment},
  author	= {Murali, Aditya and Alapatt, Deepak and Mascagni, Pietro
		  and Vardazaryan, Armine and Garcia, Alain and Okamoto,
		  Nariaki and Mutter, Didier and Padoy, Nicolas},
  journal	= {IEEE Transactions on Medical Imaging},
  volume	= {43},
  number	= {3},
  pages		= {1247--1258},
  year		= {2023},
  publisher	= {IEEE}
}

@Article{	  nwoye2023cholectriplet,
  author	= {Nwoye, C. I. and Alapatt, D. and Yu, T. and Vardazaryan,
		  A. and Xia, F. and Zhao, Z. and Xia, T. and Jia, F. and
		  Yang, Y. and Wang, H. and others},
  title		= {{CholecTriplet2021}: A Benchmark Challenge for Surgical
		  Action Triplet Recognition},
  journal	= {Medical Image Analysis},
  volume	= {86},
  pages		= {102803},
  year		= {2023}
}

@InProceedings{	  nwoye2020recognition,
  author	= {Nwoye, C. I. and Gonzalez, C. and Yu, T. and Mascagni, P.
		  and Mutter, D. and Marescaux, J. and Padoy, N.},
  title		= {Recognition of Instrument-Tissue Interactions in
		  Endoscopic Videos via Action Triplets},
  booktitle	= miccai,
  pages		= {364--374},
  year		= {2020}
}

@Article{	  nwoye2022rendezvous,
  author	= {Nwoye, C. I. and Yu, T. and Gonzalez, C. and Seeliger, B.
		  and Mascagni, P. and Mutter, D. and Marescaux, J. and
		  Padoy, N.},
  title		= {Rendezvous: Attention Mechanisms for the Recognition of
		  Surgical Action Triplets in Endoscopic Videos},
  journal	= {Medical Image Analysis},
  volume	= {78},
  pages		= {102433},
  year		= {2022}
}

@Article{	  nwoye2023cholectriplet2022,
  author	= {Nwoye, C. I. and Yu, T. and Sharma, S. and Murali, A. and
		  Alapatt, D. and Vardazaryan, A. and Yuan, K. and Hajek, J.
		  and Reiter, W. and Yamlahi, A. and others},
  title		= {{CholecTriplet2022}: Show Me a Tool and Tell Me the
		  Triplet -- An Endoscopic Vision Challenge for Surgical
		  Action Triplet Detection},
  journal	= {Medical Image Analysis},
  volume	= {89},
  pages		= {102888},
  year		= {2023}
}

@article{ozsoy2024specialized,
  title={Specialized foundation models for intelligent operating rooms},
  author={{\"O}zsoy, Ege and Pellegrini, Chantal and Bani-Harouni, David and Yuan, Kun and Keicher, Matthias and Navab, Nassir},
  journal={npj Digital Medicine},
  year={2026}
}

@inproceedings{ozsoy2024oracle,
  title={{ORacle}: Large vision-language models for knowledge-guided holistic {OR} domain modeling},
  author={{\"O}zsoy, Ege and Pellegrini, Chantal and Keicher, Matthias and Navab, Nassir},
  booktitle=miccai,
  pages={455--465},
  year={2024},
  organization={Springer}
}

@InProceedings{	  rodin2024action,
  author	= {Rodin, Ivan and Furnari, Antonino and Min, Kyle and
		  Tripathi, Subarna and Farinella, Giovanni Maria},
  title		= {Action Scene Graphs for Long-form Understanding of
		  Egocentric Videos},
  booktitle	= cvpr,
  pages		= {18622--18632},
  year		= {2024}
}

@Article{	  srivastav2018mvor,
  author	= {Srivastav, V. and Issenhuth, T. and Kadkhodamohammadi, A.
		  and de Mathelin, M. and Gangi, A. and Padoy, N.},
  title		= {{MVOR}: A Multi-view {RGB-D} Operating Room Dataset for
		  {2D} and {3D} Human Pose Estimation},
  journal	= {arXiv preprint arXiv:1808.08180},
  year		= {2018}
}

@InProceedings{	  tripathi2019compact,
  author	= {Tripathi, S. and Sridhar, S. N. and Sundaresan, S. and
		  Tang, H.},
  title		= {Compact Scene Graphs for Layout Composition and Patch
		  Retrieval},
  booktitle	= cvprw,
  year		= {2019}
}

@InProceedings{	  wang2023dynamic,
  author	= {Wang, H. and Jin, Y. and Zhu, L.},
  title		= {Dynamic Interactive Relation Capturing via Scene Graph
		  Learning for Robotic Surgical Report Generation},
  booktitle	= icra,
  pages		= {2702--2709},
  year		= {2023}
}

@Article{	  yuan2024advancing,
  author	= {Yuan, Kun and Kattel, M. and Lavanchy, J. L. and Navab,
		  Nassir and Srivastav, V. and Padoy, Nicolas},
  title		= {Advancing Surgical {VQA} with Scene Graph Knowledge},
  journal	= {International Journal of Computer Assisted Radiology and
		  Surgery},
  volume	= {19},
  pages		= {1409--1417},
  year		= {2024}
}

@InProceedings{	  ozsoy2025pix2sg,
  author	= {{\"O}zsoy, Ege and Holm, Felix and Pellegrini, Chantal and
		  Czempiel, Tobias and Saleh, Mahdi and Navab, Nassir and
		  Busam, Benjamin},
  title		= {Location-Free Scene Graph Generation},
  booktitle	= cvprw,
  year		= {2025}
}

@Article{	  henriques2025decoding,
  author	= {Henriques, Angelo and Hoxha, Korab and Zapp, Daniel and
		  Issa, Peter C. and Navab, Nassir and Nasseri, M. Ali},
  title		= {Decoding the Surgical Scene: A Scoping Review of Scene
		  Graphs in Surgery},
  journal	= {arXiv preprint arXiv:2509.20941},
  year		= {2025}
}

@Article{	  kuhn1955hungarian,
  title		= {The {Hungarian} method for the assignment problem},
  author	= {Kuhn, Harold W.},
  journal	= {Naval Research Logistics Quarterly},
  volume	= {2},
  number	= {1-2},
  pages		= {83--97},
  year		= {1955},
  publisher	= {Wiley}
}

@InProceedings{	  carion2020detr,
  title		= {End-to-end object detection with transformers},
  author	= {Carion, Nicolas and Massa, Francisco and Synnaeve, Gabriel
		  and Usunier, Nicolas and Kirillov, Alexander and Zagoruyko,
		  Sergey},
  booktitle	= eccv,
  pages		= {213--229},
  year		= {2020},
  organization	= {Springer}
}

@InProceedings{	  kim2025nlvsgg,
  author	= {Kim, Kibum and Yoon, Kanghoon and In, Yeonjun and Jeon,
		  Jaehyeong and Moon, Jinyoung and Kim, Donghyun and Park,
		  Chanyoung},
  title		= {Weakly Supervised Video Scene Graph Generation via Natural
		  Language Supervision},
  booktitle	= iclr,
  year		= {2025}
}

@Article{	  lee2026pals,
  author	= {Kang, Minseok and Lee, Minhyeok and Kim, Minjung and Lee,
		  Jungho and Kim, Donghyeong and Woo, Sungmin and Jeon,
		  Inseok and Lee, Sangyoun},
  title		= {Revisiting Weakly-Supervised Video Scene Graph Generation
		  via Pair Affinity Learning},
  journal	= {arXiv preprint arXiv:2603.21559},
  year		= {2026}
}

@InProceedings{	  yao2021visualds,
  author	= {Yao, Yuan and Zhang, Ao and Han, Xu and Li, Mengdi and Weber, Cornelius and Liu, Zhiyuan and Wermter, Stefan and Sun, Maosong},
  title		= {Visual Distant Supervision for Scene Graph Generation},
  booktitle	= iccv,
  pages={15796--15806},
  year		= {2021}
}

@InProceedings{	  wuetal2024_ovvsgg,
  author	= {Wu, Ziyue and Gao, Junyu and Xu, Changsheng},
  title		= {Open-Vocabulary Video Scene Graph Generation via
		  Union-aware Semantic Alignment},
  booktitle	= {Proceedings of the 32nd ACM International Conference on
		  Multimedia},
  pages		= {8566--8575},
  year		= {2024}
}

@InProceedings{	  zhong2021sgnls,
  author	= {Zhong, Yiwu and Shi, Jing and Yang, Jianwei and Xu,
		  Chenliang and Li, Yin},
  title		= {Learning to Generate Scene Graph from Natural Language
		  Supervision},
  booktitle	= iccv,
  pages		= {1823--1834},
  year		= {2021}
}

@InProceedings{	  ye2021lsws,
  author	= {Ye, Keren and Kovashka, Adriana},
  title		= {Linguistic Structures as Weak Supervision for Visual Scene
		  Graph Generation},
  booktitle	= cvpr,
  pages		= {8289--8299},
  year		= {2021}
}

@InProceedings{	  zhang2023vs3,
  author	= {Zhang, Yong and Pan, Yingwei and Yao, Ting and Huang, Rui
		  and Mei, Tao and Chen, Chang-Wen},
  title		= {Learning to Generate Language-Supervised and
		  Open-Vocabulary Scene Graph Using Pre-Trained
		  Visual-Semantic Space},
  booktitle	= cvpr,
  pages		= {2915--2924},
  year		= {2023}
}

@InProceedings{	  kim2024llm4sgg,
  author	= {Kim, Kibum and Yoon, Kanghoon and Jeon, Jaehyeong and In,
		  Yeonjun and Moon, Jinyoung and Kim, Donghyun and Park,
		  Chanyoung},
  title		= {{LLM4SGG}: Large Language Models for Weakly Supervised
		  Scene Graph Generation},
  booktitle	= cvpr,
  pages		= {28306--28316},
  year		= {2024}
}

@InProceedings{	  chen2023pla,
  author	= {Chen, Siqi and Xiao, Jun and Chen, Long},
  title		= {Video Scene Graph Generation from Single-Frame Weak
		  Supervision},
  booktitle	= iclr,
  year		= {2023}
}

\clearpage
\appendix
\setcounter{section}{0}
\setcounter{figure}{0}
\setcounter{table}{0}
\setcounter{equation}{0}

\renewcommand{\thesection}{\Alph{section}}
\renewcommand{\thefigure}{S\arabic{figure}}
\renewcommand{\thetable}{S\arabic{table}}
\renewcommand{\theequation}{S\arabic{equation}}

\section*{Appendix}

This supplementary material provides implementation details, ablations, and qualitative analyses supporting the design choices described in the main text. Section~\ref{sec:segcache} (Segmentation Cache) details our dual-resolution SAM3 inference strategy, the prompt selection process underlying the recall-first design, and a prompt ablation justifying the choice of general categories over fine-grained surgical terminology. Section~\ref{sec:3dcache} (3D Scene Representation Cache) presents qualitative comparisons between Pi3 zero-shot reconstructions and the calibrated depth sensor outputs used by prior 4D-OR methods, demonstrating that RGB-only reconstruction provides sufficient geometric fidelity for our pipeline. Section~\ref{sec:fusion} (Multi-View Instance Fusion) formalizes the category-constrained 3D IoU clustering used to merge cross-camera masks into unified instances, including the adaptive voxelization scheme that handles Pi3's scale ambiguity. Section~\ref{sec:visfeat} (Visual Feature Extraction) describes the hybrid-resolution DINOv3 extraction strategy that enables high-fidelity features for small surgical instruments. Section~\ref{sec:arch} (Architectural Details) provides complete specifications of the graph transformer, including positional encoding, edge-aware attention, Hungarian matching cost construction, and complete loss formulation. Section~\ref{sec:control} (Control Study) reports detection statistics for the irrelevant-prompt control experiment that validates the unsupervised context augmentation result. Together, these sections expand on design decisions, hyperparameter choices, and validation experiments that space constraints prevented us from including in the main text.

\section{Segmentation Cache}
\label{sec:segcache}

\subsection{Dual-Resolution Segmentation Strategy}

SAM3's open-vocabulary segmentation performance degrades significantly for small objects when processing low-resolution images. In our preliminary experiments, running SAM3 on downsampled images ($392 \times 518$) with the prompt \textit{tool} yielded 0\% recall for surgical instruments because the model could not detect objects occupying fewer than $\sim$50 pixels. This motivated a dual-resolution inference strategy that is critical to achieving our reported 100\% tool recall.

\textbf{Resolution selection.} We partition prompts by expected object size:
\begin{itemize}
    \item \textbf{Low-resolution} ($392 \times 518$): Large entities (person, medical table, anesthesia equipment). These occupy substantial image area and are reliably detected at reduced resolution.
    \item \textbf{High-resolution} ($1536 \times 2048$): Small entities (tool, hand). Processing at native resolution provides the pixel density required for SAM3 to recognize fine-grained structures.
\end{itemize}

\textbf{Confidence thresholds.} We apply category-specific confidence thresholds to balance precision and recall:
\begin{itemize}
    \item Standard threshold (0.3): person, table, anesthesia, hand
    \item Permissive threshold (0.01): tool. SAM3 produces low-confidence detections for surgical instruments even at high resolution, so we accept nearly all detections and rely on downstream fusion and Hungarian matching to filter false positives.
\end{itemize}

\textbf{Prompt engineering.} The specific prompt text significantly impacts detection quality. For instruments, we use \textit{tool in hand} rather than \textit{surgical instrument}, as surgical instrument detects nearly every object in the surgical scene rather than the tools we are interested in. This is not a failure mode of the model as these objects are technically related to the surgery (\textit{e.g.,} anesthesia machine, monitor, tables), but these objects are either already segmented by other prompts or not necessary for the downstream goals. Table~\ref{tab:sam3_combined} shows the performance for final prompts while Table~\ref{tab:prompt_ablation} shows detailed performance for prompts from exploratory experiments. 

\textbf{Output normalization.} All masks are resized to a common output resolution ($392 \times 518$) using nearest-neighbor interpolation to preserve binary mask boundaries. This ensures consistent downstream processing regardless of the inference resolution.

\subsection{Quantitative Analysis}

Table~\ref{tab:sam3_combined} reports segmentation recall across dataset train and test splits. Recall is computed as the ratio of fused SAM3 instances to ground truth object counts. Person, anesthesia, and tool achieve 100\% recall across splits, validating our high-recall segmentation strategy. The over-segmentation inherent in our approach introduces false positives that are filtered during semi-supervised Hungarian matching, enabling the graph transformer to focus on relational reasoning without concern for missed entities. However, unlike other prompts that achieve 100\% recall seen in Table~\ref{tab:prompt_ablation}, the final prompts do not encroach on each other's areas of responsibility. For example, the \textit{tool} prompt does not segment anesthesia machines and the \textit{Table} prompt does not label other types of surgical equipment. 

\begin{table}[t]
\centering
\caption{SAM3 configuration and resulting segmentation recall. Left: prompt configuration with per-category resolution and confidence thresholds. Right: recall across dataset splits (over-segmentation handled by Hungarian matching; hand has no GT labels).}
\label{tab:sam3_combined}
\small
\begin{minipage}{0.58\linewidth}
\centering
\begin{tabular}{@{}llcc@{}}
\toprule
Category & Prompt Text & Res. & Conf. \\
\midrule
Person     & ``person''               & Low  & 0.3  \\
Table      & ``medical table''        & Low  & 0.3  \\
Anesthesia & ``anesthesia equipment'' & Low  & 0.3  \\
Tool       & ``tool in hand''         & High & 0.01 \\
Hand       & ``hand''                 & High & 0.3  \\
\bottomrule
\end{tabular}

\end{minipage}\hfill
\begin{minipage}{0.38\linewidth}
\centering
\begin{tabular}{@{}lcc@{}}
\toprule
Category & Train & Test \\
\midrule
Person     & 100\% & 100\% \\
Table      & 98\%  & 97\%  \\
Anesthesia & 100\% & 100\% \\
Tool       & 100\% & 100\% \\
Hand       & ---   & ---   \\
\bottomrule
\end{tabular}

\end{minipage}
\end{table}

\subsection{Prompt Ablation Study}
\label{sec:prompt_ablation}

To validate our choice of general prompts over fine-grained alternatives, we conducted an ablation study comparing detection behavior across different prompt formulations on the training set. Table~\ref{tab:prompt_ablation} reports the results grouped by target category.

\begin{table}[t]
\centering
\caption{Prompt ablation on the training set (4,179 frames $\times$ 6 cameras). Fine-grained prompts either fail to detect targets or indiscriminately label all instances.}
\label{tab:prompt_ablation}
\small
\begin{tabular}{@{}llrr@{}}
\toprule
Target & Prompt Text & Det. & Recall \\
\midrule
\multicolumn{4}{@{}l}{\textit{Person variants}} \\
\quad Patient only & ``patient'' & 14,850 & 100\%$^\dagger$ \\
\quad Staff only & ``doctor'' & 25,704 & 100\%$^\dagger$ \\
\quad Staff only & ``nurse'' & 23,502 & 100\%$^\dagger$ \\
\quad Staff only & ``surgeon'' & 35,637 & 100\%$^\dagger$ \\
\quad Staff only & ``assistant'' & 11 & 0\% \\
\midrule
\multicolumn{4}{@{}l}{\textit{Table variants}} \\
\quad Operating & ``operating table'' & 4,687 & 100\%$^\dagger$ \\
\quad Secondary & ``secondary table'' & 44 & 2\% \\
\quad Instrument & ``instrument table'' & 0 & 0\% \\
\quad Instrument & ``tool table'' & 2,445 & 100\%$^\dagger$ \\
\midrule
\multicolumn{4}{@{}l}{\textit{Tool variants}} \\
\quad All tools & ``surgical tool'' & 627 & 30\% \\
\quad All tools & ``medical instrument'' & 231,001 & 100\%$^\dagger$ \\

\quad All tools & ``saw'' & 0 & 0\% \\
\quad All tools & ``needle'' & 20 & 1\% \\
\quad All tools & ``scalpel'' & 0 & 0\% \\
\quad All tools & ``forceps'' & 2 & 0\% \\

\bottomrule
\end{tabular}
\\[2pt]
\raggedright\scriptsize $^\dagger$Achieves 100\% recall but with severe category confusion (see \ref{sec:prompt_ablation}).
\end{table}

The results reveal two failure modes for fine-grained prompts:

\textbf{Category confusion.} Role-specific person prompts (``doctor'', ``nurse'') achieve 100\% recall but with 5--7$\times$ over-detection because SAM3 labels every person in the scene with each role, including the patient. Similarly, \textit{operating table} detects all tables regardless of type, and \textit{medical instrument} produces 231,001 detections by labeling nearly every object in every frame. This confusion would propagate incorrect category hints to the downstream graph transformer, degrading entity disambiguation.

\textbf{Detection failure.} Specific prompts often fail entirely: \textit{instrument table} achieves 0\% recall, \textit{secondary table} only 2\%, and tool-specific prompts like \textit{saw}, \textit{scalpel}, and \textit{forceps} detect nothing. These prompts are narrow, requiring significantly increased resolution and cropping, which is a performance bottleneck. Moreover, using fine-grained prompts would degrade the adaptability of our pipeline because new surgical settings would require intricate prompt engineering for new entities. Keeping the prompts general allows for new entities to be introduced under general prompts that have high recall without complication while letting the graph transformer shoulder the challenging portion of the task. 

Our general prompt strategy avoids both failure modes. The \textit{person}, \textit{medical table}, and \textit{tool in hand} prompts achieve 100\% recall without category confusion. The downstream graph transformer then performs fine-grained disambiguation using visual features and spatial reasoning.

\subsection{Qualitative Analysis}

\begin{figure*}[h!]
\centering
\includegraphics[width=0.9\textwidth]{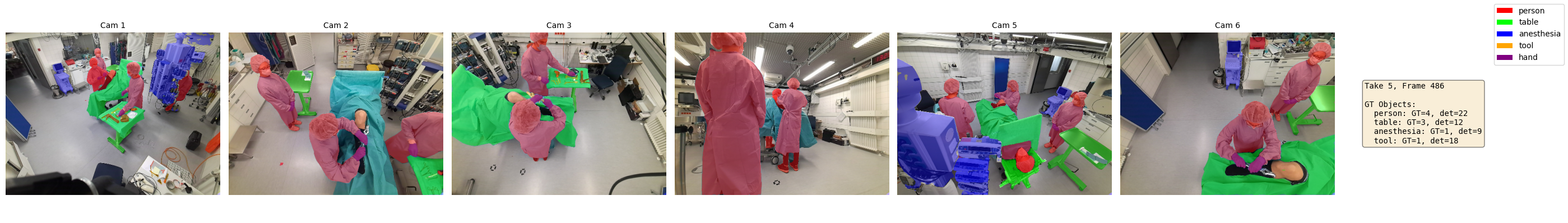} \\
\includegraphics[width=0.9\textwidth]{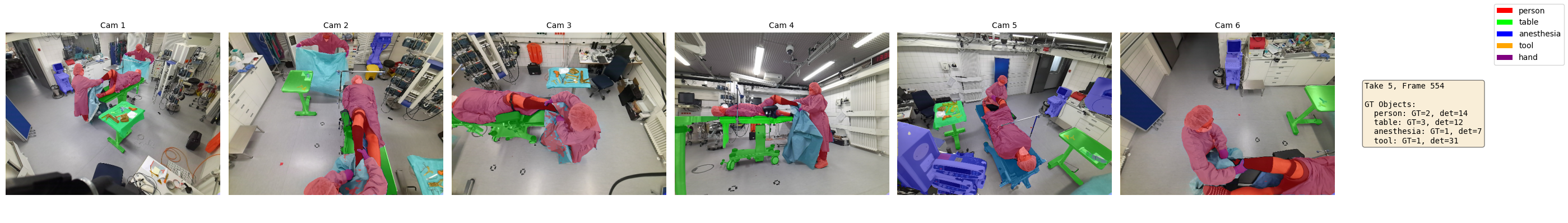} \\
\includegraphics[width=0.9\textwidth]{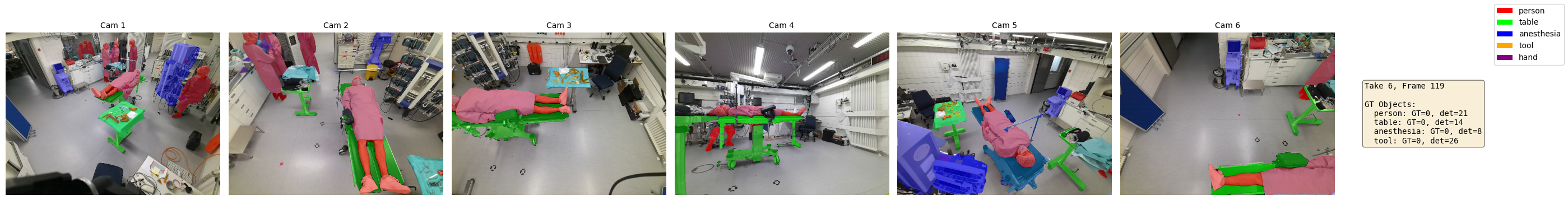} \\
\includegraphics[width=0.9\textwidth]{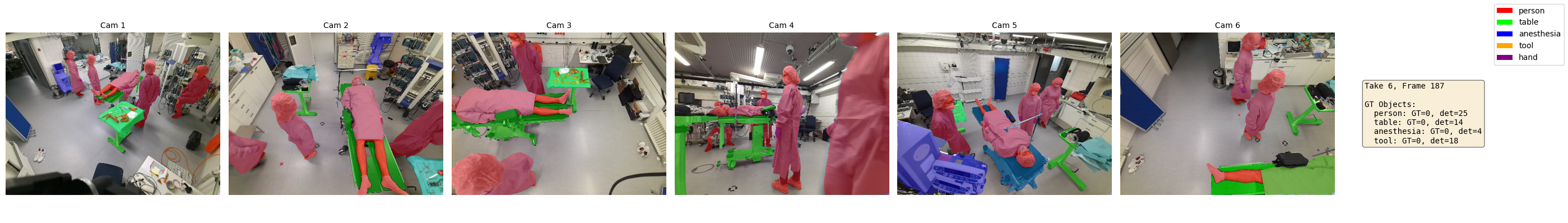} \\
\includegraphics[width=0.9\textwidth]{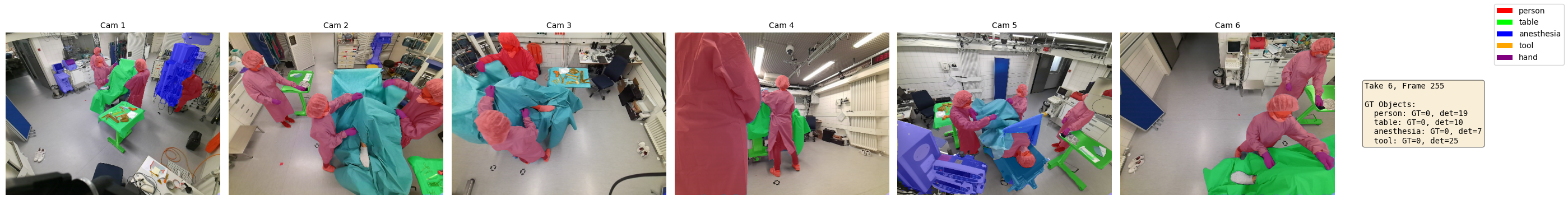} \\
\includegraphics[width=0.9\textwidth]{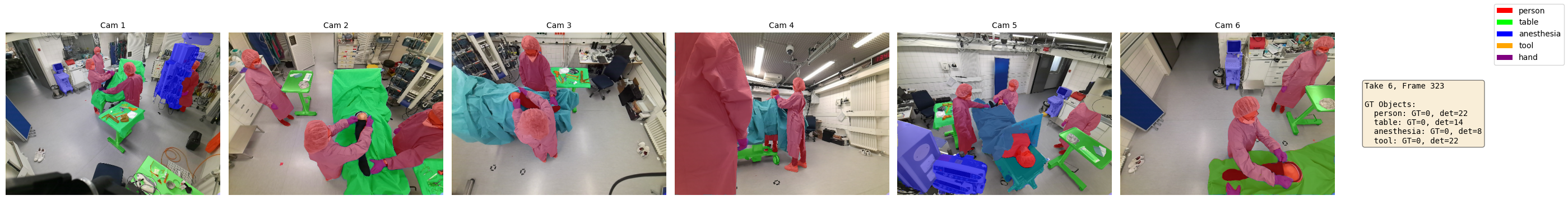} \\
\includegraphics[width=0.9\textwidth]{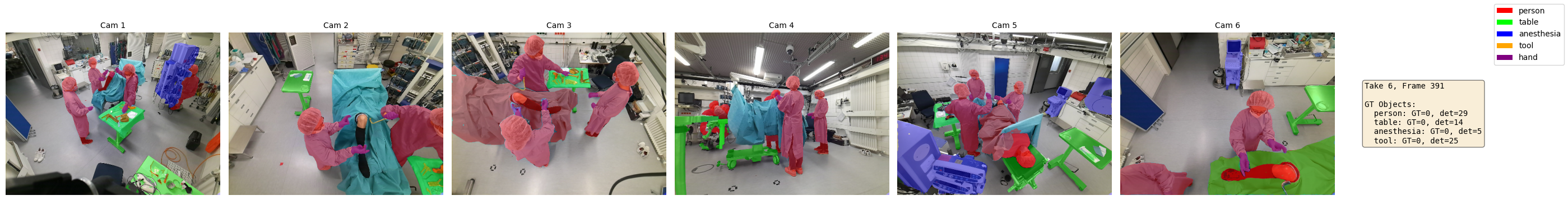} \\
\includegraphics[width=0.9\textwidth]{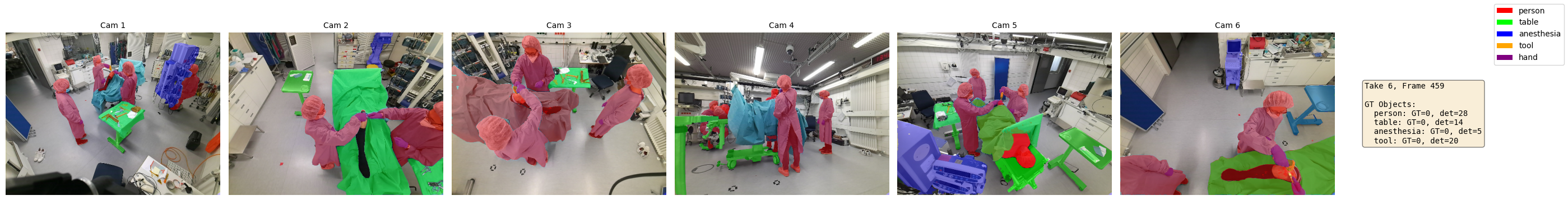} \\

\caption{Qualitative segmentation examples from the train set. Each visualization shows the original camera views with instance masks overlaid. Colors indicate SAM3 categories. Examples demonstrate high recall across diverse surgical contexts.}
\label{fig:segmentation_examples}
\end{figure*}

Figure~\ref{fig:segmentation_examples} shows representative examples of our cached segmentation results. Each row displays the original camera views overlaid with all detected instances (color-coded by category). Crucially, if SAM3 misses an object in an image, it can be recovered in the 3D fused instances as long as one of the 6 images in the frame has that correctly segmented object. This guards against another potential failure mode. The fusion strategy is described in detail in section~\ref{sec:fusion}.

\section{3D Scene Representation Cache}
\label{sec:3dcache}

\begin{figure*}
\centering
\begin{tabular}{cc}
\textbf{Pi3 (Ours)} & \textbf{4D-OR Depth Sensor} \\
\includegraphics[width=0.45\textwidth]{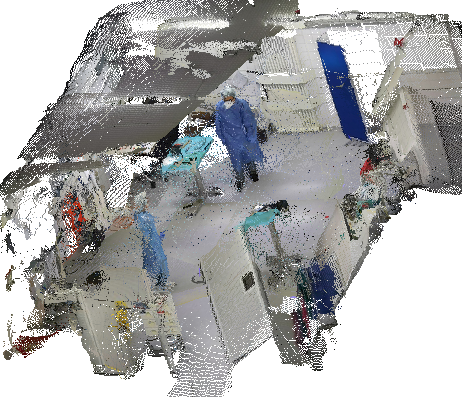} &
\includegraphics[width=0.45\textwidth]{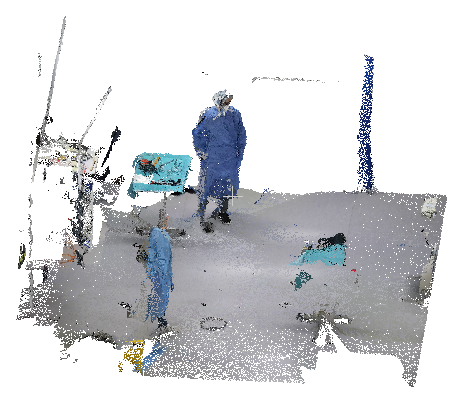} \\ [20pt]
\includegraphics[width=0.45\textwidth]{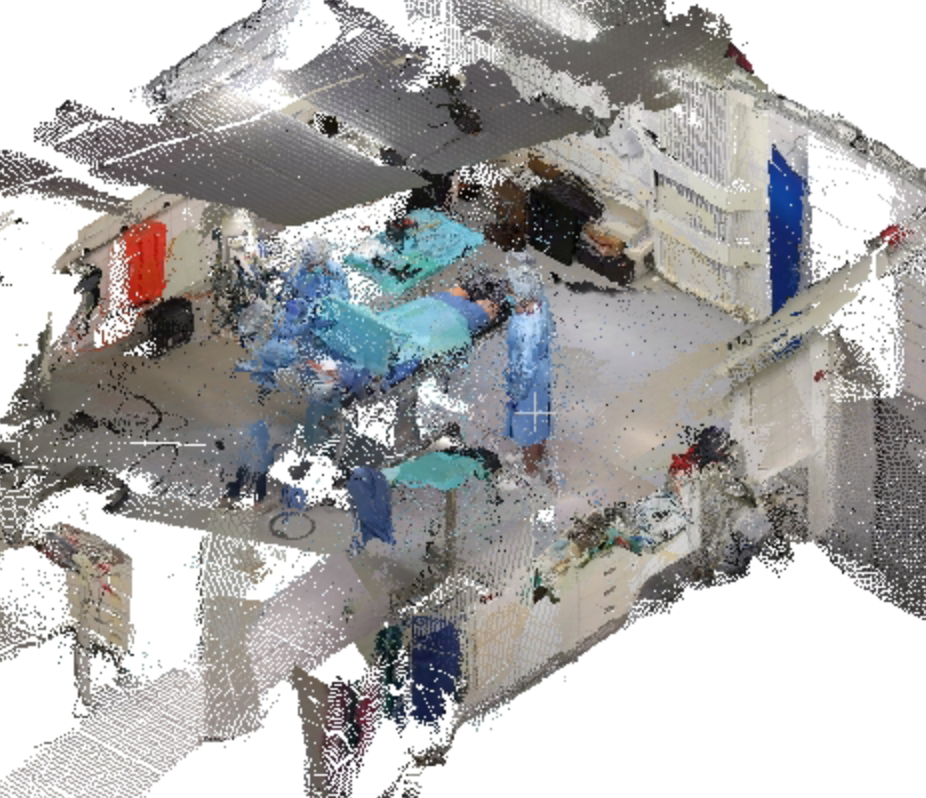} &
\includegraphics[width=0.45\textwidth]{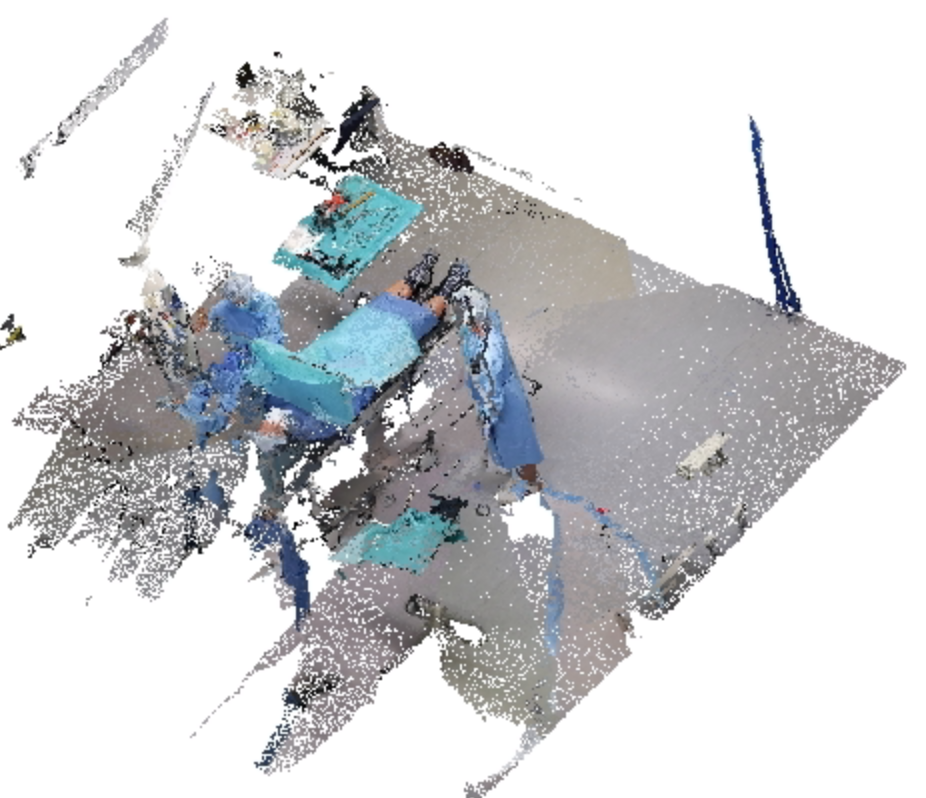} \\ [20pt]
\includegraphics[width=0.45\textwidth]{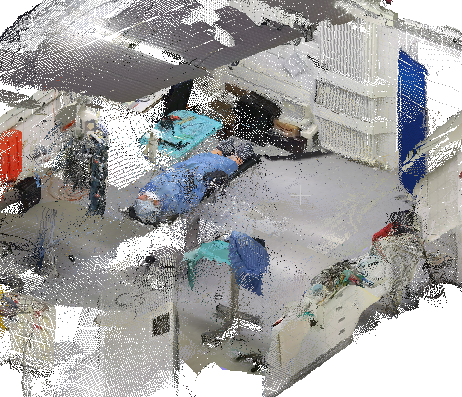} &
\includegraphics[width=0.45\textwidth]{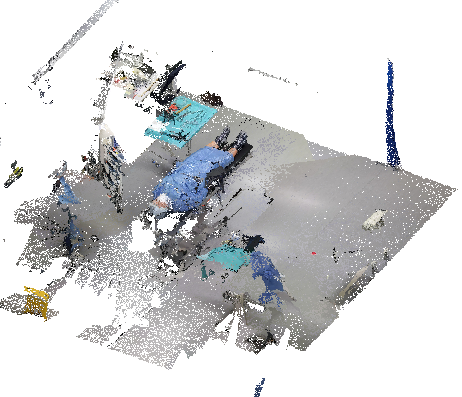} 
\end{tabular}
\vspace{0.5cm}
\caption{Qualitative comparison of Pi3 zero-shot reconstruction (left) versus 4D-OR depth sensor point clouds (right) on matched frames. Pi3 produces significantly denser point clouds with finer geometric detail, enabling robust centroid extraction for diverse entity types.}
\label{fig:pi3_vs_4dor}

\end{figure*}

We compare Pi3's zero-shot multi-view 3D reconstruction against 4D-OR's depth sensor point clouds in Figure~\ref{fig:pi3_vs_4dor}. Pi3 consistently produces denser, more detailed point clouds that enable accurate centroid extraction for any entity in the scene. Although Pi3 point clouds exhibit slightly higher noise levels compared to depth sensors, this is not a concern for our pipeline: we filter point clouds by extracting centroids only within segmentation masks provided by SAM3, effectively removing spurious points. Increased density is critical for capturing fine-grained structures, such as surgical instruments, anesthesia equipment, and patient anatomy that may be poorly represented in sparse depth sensor outputs. Additionally, not only do we not require depth sensors, but because Pi3 outputs relative poses for all cameras, no calibration is required, and the pipeline is adaptable to any number of cameras. 

\section{Fusion}
\label{sec:fusion}

Multi-view instance fusion resolves the many-to-many correspondence between 2D segmentation masks across cameras by lifting them into 3D and clustering via spatial overlap. This process produces a unified set of fused instances that represent physical entities in the scene, regardless of which cameras observe them. The fusion algorithm addresses two key challenges: (1) SAM3's category labels are inconsistent across views (\textit{e.g.,} an operating table may be labeled \textit{table} in one camera and \textit{anesthesia} in another due to viewing angle or an object may be segmented in one view while being missed in another), and (2) Pi3's reconstructed depth is relative and scale-ambiguous, precluding the use of fixed metric thresholds.

\subsection{Point Cloud Extraction}

For each 2D mask $m$ in camera $c$, we extract 3D points from the Pi3 pointmap $\mathbf{P}_c \in \mathbb{R}^{H \times W \times 3}$ and transform them to world coordinates. Let $\mathbf{C}_c \in \mathbb{R}^{H \times W}$ denote the Pi3 confidence map. The valid point set within the mask is:
\begin{equation}
\mathcal{V} = \{(u, v) \mid m(u,v) = 1 \text{ and } \mathbf{C}_c(u,v) > \tau_{\text{conf}}\}
\end{equation}
where $\tau_{\text{conf}} = 0.1$ is the confidence threshold. We extract local 3D points $\{\mathbf{p}_{\text{local}}^{(i)}\}_{i \in \mathcal{V}}$ from $\mathbf{P}_c$ and transform to world coordinates:
\begin{equation}
\mathbf{p}_{\text{world}}^{(i)} = \mathbf{R}_c \mathbf{p}_{\text{local}}^{(i)} + \mathbf{t}_c
\end{equation}
where $[\mathbf{R}_c | \mathbf{t}_c] \in \mathbb{R}^{4 \times 4}$ is the camera-to-world extrinsic matrix from Pi3. If $|\mathcal{V}| < n_{\min}$ (default: 10 points), the mask is discarded as spurious. If $|\mathcal{V}| > n_{\max}$ (default: 1000 points), we uniformly subsample to $n_{\max}$ for computational efficiency. The resulting point cloud $\mathcal{P}_m = \{\mathbf{p}_{\text{world}}^{(i)}\}$ and its centroid $\bar{\mathbf{p}}_m = \frac{1}{|\mathcal{P}_m|} \sum \mathbf{p}_{\text{world}}^{(i)}$ characterize the mask in 3D.

\subsection{3D IoU Computation via Voxelization}

To measure spatial overlap between two masks' point clouds $\mathcal{P}_1$ and $\mathcal{P}_2$, we compute 3D IoU via voxelization. Given a voxel size $s_v$, we discretize each point cloud into a set of occupied voxels:
\begin{equation}
\mathcal{G}(\mathcal{P}, s_v) = \left\{\left\lfloor \frac{\mathbf{p}}{s_v} \right\rfloor \mid \mathbf{p} \in \mathcal{P}\right\}
\end{equation}
where $\lfloor \cdot \rfloor$ denotes element-wise floor. The 3D IoU is then:
\begin{equation}
\text{IoU}_{3D}(\mathcal{P}_1, \mathcal{P}_2) = \frac{|\mathcal{G}(\mathcal{P}_1, s_v) \cap \mathcal{G}(\mathcal{P}_2, s_v)|}{|\mathcal{G}(\mathcal{P}_1, s_v) \cup \mathcal{G}(\mathcal{P}_2, s_v)|}
\end{equation}
This voxel-based formulation is efficient (set operations via hashing) and invariant to point cloud density. The voxel size $s_v$ is chosen adaptively to account for Pi3's scale ambiguity (see Section~\ref{sec:adaptive_scaling}).

\subsection{Category-Constrained Clustering}
\label{sec:category_constraints}

SAM3 can assign inconsistent category labels to the same physical object across views (\textit{e.g.,} a table may be labeled ``table'' in one camera and ``anesthesia'' in another due to equipment placement). To address this, we perform cross-category fusion with compatibility constraints. All SAM3 categories are partitioned into three groups:
\begin{align}
G_0 &= \{\text{person}\} \\
G_1 &= \{\text{table}, \text{anesthesia}\} \\
G_2 &= \{\text{tool}, \text{hand}\}
\end{align}
Only masks whose SAM3 categories belong to the same group may be merged. This grouping reflects domain knowledge: Group 0 (people) are always distinct entities; Group 1 (large equipment) can be confused by SAM3 due to visual similarity and spatial proximity; Group 2 (small items) includes tools held in hands. When masks from different categories within a group are merged, the final category is determined by the majority vote weighted by SAM3 confidence scores.

Let $\{m_i\}_{i=1}^N$ denote all masks across all cameras, each with category $c_i$ and group assignment $g_i = G(c_i)$. We compute the pairwise IoU matrix $\mathbf{I} \in \mathbb{R}^{N \times N}$ where:
\begin{equation}
\mathbf{I}_{ij} = \begin{cases}
\text{IoU}_{3D}(\mathcal{P}_i, \mathcal{P}_j) & \text{if } g_i = g_j \\
0 & \text{otherwise}
\end{cases}
\end{equation}
Masks are merged via greedy clustering: for each pair $(i, j)$ with $\mathbf{I}_{ij} \geq \tau_{\text{IoU}}$ (default: 0.15), we assign them to the same cluster. Clusters are relabeled to consecutive integers, resulting in the final set of fused instances.

\subsection{Adaptive Scene Scaling}
\label{sec:adaptive_scaling}

Pi3 reconstructs scenes up to an arbitrary global scale, making fixed metric voxel sizes inappropriate. To achieve scale-invariant IoU computation, we adaptively set the voxel size $s_v$ based on the scene extent. Let $\mathcal{P}_{\text{all}} = \bigcup_{i=1}^N \mathcal{P}_i$ be the union of all extracted point clouds. We compute the bounding box diagonal:
\begin{equation}
D_{\text{scene}} = \left\|\max_{\mathbf{p} \in \mathcal{P}_{\text{all}}} \mathbf{p} - \min_{\mathbf{p} \in \mathcal{P}_{\text{all}}} \mathbf{p}\right\|_2
\end{equation}
and set $s_v = 0.02 \cdot \max(D_{\text{scene}}, 0.1)$. This scales voxels to 2\% of the scene diagonal, making IoU thresholds consistent regardless of Pi3's arbitrary depth scale. In practice, $D_{\text{scene}}$ typically ranges from 2--10 meters (operating room scale), resulting in voxel sizes of 2--10 cm.

\subsection{Geometric Feature Extraction}

For each fused instance with point cloud $\mathcal{P} = \{\mathbf{p}^{(i)}\}_{i=1}^{n}$ and centroid $\bar{\mathbf{p}}$, we extract a 13-dimensional geometric feature vector:
\begin{equation}
\mathbf{g} = [\bar{\mathbf{p}}, \mathbf{b}_{\text{size}}, \boldsymbol{\sigma}, n_{\text{norm}}, c_{\text{norm}}, s_{\text{conf}}, h]
\end{equation}
where the components are:

\textbf{3D Centroid} (3 dims): $\bar{\mathbf{p}} = \frac{1}{n}\sum_{i=1}^n \mathbf{p}^{(i)} \in \mathbb{R}^3$ provides the instance's world-space location.

\textbf{Bounding Box Size} (3 dims): $\mathbf{b}_{\text{size}} = [\max_i p_x^{(i)} - \min_i p_x^{(i)}, \max_i p_y^{(i)} - \min_i p_y^{(i)}, \max_i p_z^{(i)} - \min_i p_z^{(i)}]$ captures the spatial extent in each dimension.

\textbf{Point Cloud Spread} (3 dims): $\boldsymbol{\sigma} = [\sigma_x, \sigma_y, \sigma_z]$ where $\sigma_k = \sqrt{\frac{1}{n}\sum_{i=1}^n (p_k^{(i)} - \bar{p}_k)^2}$ measures the distribution compactness per axis.

\textbf{Point Density} (1 dim): $n_{\text{norm}} = n / 1000$ is the normalized point count, indicating instance size and reconstruction quality.

\textbf{Multi-View Coverage} (1 dim): $c_{\text{norm}} = |\{\text{cameras}\}| / 6$ measures the fraction of cameras observing the instance, reflecting visibility and occlusion.

\textbf{Confidence} (1 dim): $s_{\text{conf}}$ is the average SAM3 confidence score across all constituent masks.

\textbf{Vertical Extent} (1 dim): $h = b_{size,z}$ isolates the height dimension, which is semantically meaningful in operating room contexts (\textit{e.g.,} distinguishing tables from people).

These features are stored in the cache at half-precision (2 bytes per value, 26 bytes per instance) and provide the graph transformer with rich geometric cues for spatial reasoning without requiring explicit 3D supervision.

\subsection{Qualitative Analysis}

\begin{figure}[t]
\centering
\begin{tabular}{cc}
\textbf{Instance Centroids} & \textbf{Fused Point Clouds} \\
\includegraphics[width=0.45\textwidth]{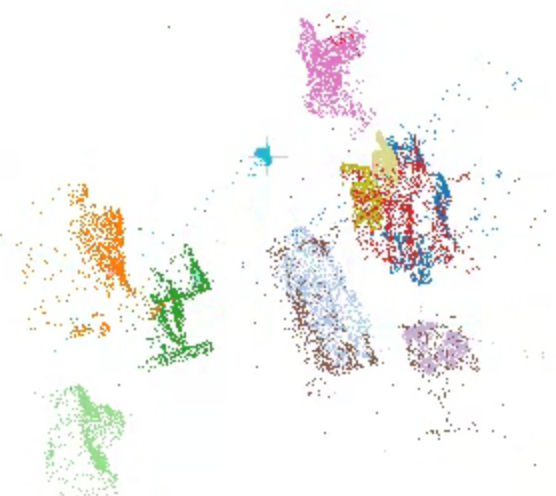} &
\includegraphics[width=0.45\textwidth]{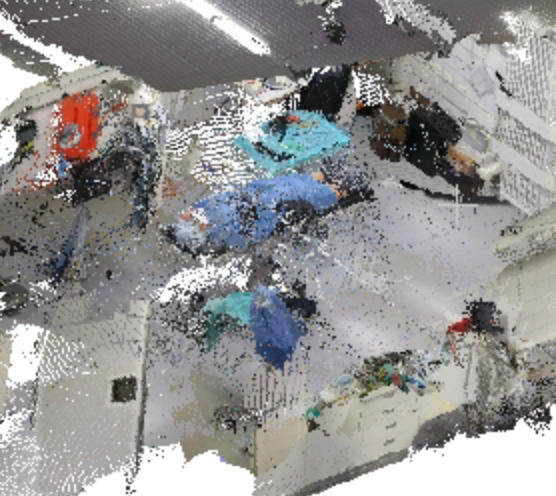} \\
\includegraphics[width=0.45\textwidth]{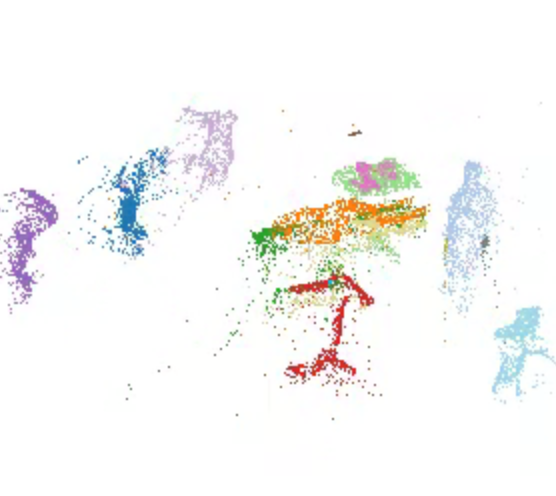} &
\includegraphics[width=0.45\textwidth]{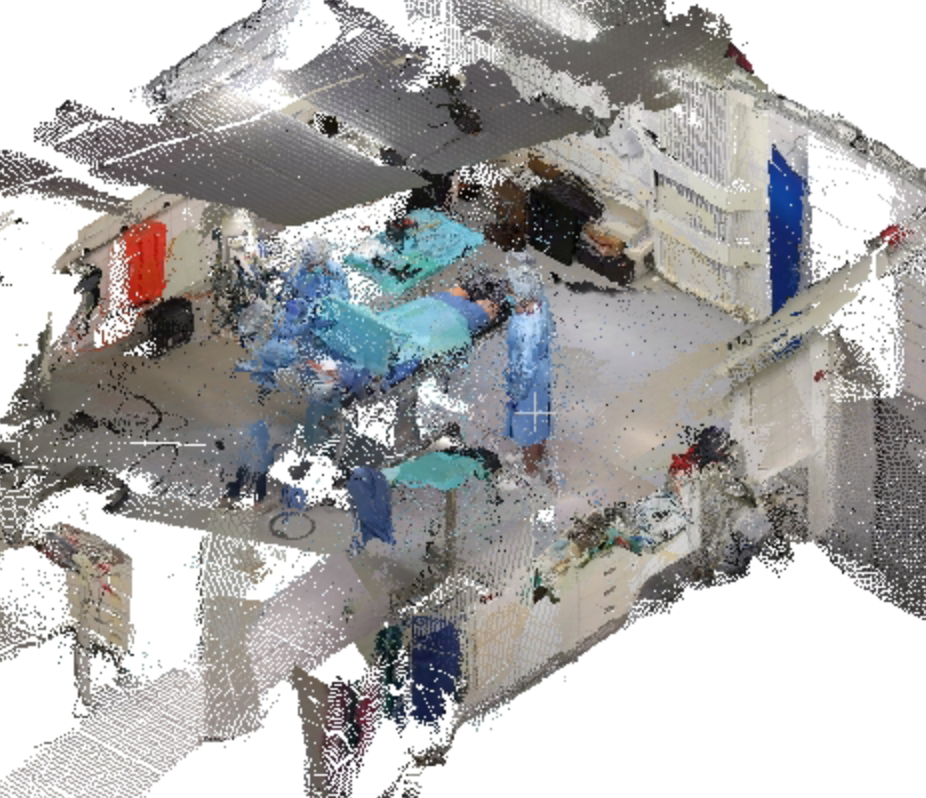}
\end{tabular}
\caption{Multi-view instance fusion results. Right: Point clouds colored by fused cluster ID after 3D IoU clustering. Left: Extracted 3D centroids (shown as spheres) for each fused instance, used as geometric features for the graph transformer. The adaptive voxelization makes fusion robust to Pi3's scale ambiguity.}
\label{fig:fusion_examples}
\end{figure}

Figure~\ref{fig:fusion_examples} shows two representative fusion results. The right column displays fused point clouds, demonstrating successful multi-view aggregation. The left column shows the corresponding 3D centroids extracted from each fused instance, which serve as geometric features for the downstream graph transformer. The adaptive voxelization and category constraints enable robust fusion despite Pi3's scale ambiguity and SAM3's cross-view category inconsistencies.

\section{Visual Feature Extraction}
\label{sec:visfeat}

We extract visual features using DINOv3 \cite{simeoni2025dinov3}, a self-supervised Vision Transformer that produces dense patch-level representations. This section details the hybrid extraction strategy and mask-to-feature pooling mechanism.

\subsection{Hybrid Extraction Strategy}

Surgical instruments and hands occupy only a small fraction of the image area in standard camera views. At the low-resolution input used for efficient processing ($392 \times 518$), a typical tool mask covers fewer than 50 patches after the ViT's $16 \times 16$ patch embedding, providing insufficient spatial resolution for discriminative feature extraction. To address this, we employ a category-aware hybrid strategy:

\textbf{Large-entity extraction.} For categories with sufficient spatial extent (person, table, anesthesia), we extract features from the standard low-resolution input. The full image is processed through DINOv3, yielding a patch grid of $24 \times 32$ tokens. Features are then pooled within each instance's mask region.

\textbf{Small-entity extraction.} For spatially compact categories (tool, hand), we crop the corresponding region from the original high-resolution image ($1536 \times 2048$) using the mask's bounding box with 20\% padding on each side. The crop is resized to $518 \times 518$ (the maximum resolution supported by DINOv3's positional embeddings) using bilinear interpolation, while the mask is resized using nearest-neighbor interpolation to preserve binary values. This yields a $32 \times 32$ patch grid dedicated entirely to the small object, providing approximately $100\times$ more patches than would be available from the low-resolution full-frame extraction.

\subsection{Mask-to-Patch Pooling}

Given an image processed by DINOv3, we obtain normalized patch tokens $\mathbf{F} \in \mathbb{R}^{h \times w \times d}$ where $h \times w$ is the patch grid size and $d = 1024$ for ViT-L. To pool features for a mask $M \in \{0, 1\}^{H \times W}$ at image resolution:

\begin{enumerate}
    \item \textbf{Mask downsampling}: Bilinearly interpolate $M$ to patch resolution $(h, w)$ and threshold at 0.5:
    \begin{equation}
        M_{\text{patch}} = \mathbf{1}\left[\text{Interpolate}(M, (h, w)) > 0.5\right]
    \end{equation}
    This soft-to-hard conversion ensures patches with majority mask coverage are included.

    \item \textbf{Feature pooling}: Mean-pool over all patches within the thresholded mask:
    \begin{equation}
        \mathbf{v} = \frac{1}{|M_{\text{patch}}|} \sum_{(i,j): M_{\text{patch}}(i,j) = 1} \mathbf{F}_{i,j}
    \end{equation}
    If no patches satisfy the threshold (empty mask), a zero vector is returned.
\end{enumerate}

\subsection{Multi-View Aggregation}

For instances visible in multiple cameras, we extract features independently from each view and aggregate via mean pooling:
\begin{equation}
    \mathbf{v}_{\text{instance}} = \frac{1}{|\mathcal{C}|} \sum_{c \in \mathcal{C}} \mathbf{v}_c
\end{equation}
where $\mathcal{C}$ is the set of cameras observing the instance. We evaluated max-pooling and attention-weighted aggregation but found mean pooling to perform best, likely because it preserves information from partially occluded views rather than being dominated by the most confident view.

\subsection{Preprocessing and Storage}

\textbf{Normalization.} Images are normalized using ImageNet statistics (mean $= [0.485, 0.456, 0.406]$, std $= [0.229, 0.224, 0.225]$) before DINOv3 inference.

\textbf{DINOv3 output.} We use the normalized patch tokens (\texttt{x\_norm\_patchtokens}) from DINOv3's forward pass, excluding the CLS token and any storage tokens. This provides spatially-aligned features suitable for mask pooling.

\textbf{Storage format.} Visual features are stored at half-precision (float16) to reduce cache size. At 1024 dimensions $\times$ 2 bytes = 2 KB per instance, combined with geometric features (26 bytes) and category hints (10 bytes), each instance requires approximately 2 KB of storage. A typical frame with 15--25 instances occupies 30--50 KB, making the complete dataset cache (6,734 frames) approximately 300 MB, three orders of magnitude smaller than the raw images it replaces.

\section{Architectural Details}
\label{sec:arch}

\begin{table}[h!]
\centering
\caption{SAGE-OR pipeline overview. Only the graph transformer is trainable; all other stages use frozen foundation models and run offline.}
\label{tab:pipeline_summary}
\small
\begin{tabular}{@{}l l l c@{}}
\toprule
Stage & Input & Output & Trainable \\
\midrule
Pi3 \cite{wang2026pi3} & Multi-view RGB & Pointmaps, poses & \texttimes \\
SAM3 \cite{carion2025sam3} & RGB + prompts & Instance masks & \texttimes \\
DINOv3 \cite{simeoni2025dinov3} & Masked crops & Visual features & \texttimes \\
Graph Transformer & Cached features & Scene graph & \checkmark \\
\bottomrule
\end{tabular}
\end{table}

\subsection{3D Positional Encoding}

To encode where each entity is located in the physical operating room, we add a 3D sinusoidal positional encoding of the world-space centroid $\mathbf{p} = (x, y, z)$. For each spatial dimension $k \in \{x, y, z\}$ and frequency index $j = 0, \ldots, D/6 - 1$, we compute:
\begin{equation}
    \text{PE}_{k,2j} = \sin\left(\frac{p_k}{10000^{2j/D}}\right), \quad
    \text{PE}_{k,2j+1} = \cos\left(\frac{p_k}{10000^{2j/D}}\right),
\end{equation}
where $D$ is the encoding dimension. The concatenated encoding across all three dimensions is projected to the model's hidden dimension via a learned linear layer and added to the input embedding.

\subsection{Edge-Aware Attention Bias}

The attention logits in each graph transformer layer are augmented with a learned spatial bias. For nodes $i$ and $j$ with world-space centroids $\mathbf{p}_i$ and $\mathbf{p}_j$, we compute a pairwise distance feature $d_{ij} = \|\mathbf{p}_i - \mathbf{p}_j\|_2$ and project it through a learned linear layer to produce a scalar bias added to the attention logits before the softmax:
\begin{equation}
    \text{Attn}(i, j) = \text{softmax}\left(\frac{\mathbf{q}_i^\top \mathbf{k}_j}{\sqrt{d_k}} + \phi(d_{ij})\right),
\label{eq:edge_attn}
\end{equation}
where $\phi$ is a learned projection from the scalar distance to a per-head bias.

\subsection{Graph Transformer Encoder}

The encoder processes cached multi-modal features through $L=4$ transformer layers with $H=6$ attention heads and hidden dimension $d=384$.

\textbf{Input Projection.} Cached features are concatenated and projected:
\begin{equation}
\mathbf{x}_i^{(0)} = \text{Dropout}(\text{LN}(\mathbf{W}_{\text{in}} [\mathbf{v}_i, \mathbf{g}_i, \mathbf{c}_i] + \mathbf{b}_{\text{in}}), p=0.3)
\end{equation}
where $\mathbf{v}_i \in \mathbb{R}^{1024}$ (visual), $\mathbf{g}_i \in \mathbb{R}^3$ (3D centroid from the 13-dim cached geometric features, see Fusion section), $\mathbf{c}_i \in \mathbb{R}^5$ (category hint), and $\mathbf{W}_{\text{in}} \in \mathbb{R}^{384 \times 1032}$ (397,440 parameters). Positional encoding $\text{PE}(\mathbf{p}_i) \in \mathbb{R}^{384}$ is added: $\mathbf{x}_i^{(0)} \leftarrow \mathbf{x}_i^{(0)} + \text{PE}(\mathbf{p}_i)$.

\textbf{Transformer Layers.} Each layer $\ell = 1, \ldots, L$ applies pre-norm edge-aware attention and feed-forward:
\begin{align}
\mathbf{x}_i^{(\ell)} &= \mathbf{x}_i^{(\ell-1)} + \text{EdgeAttn}(\text{LN}(\mathbf{x}_i^{(\ell-1)}), \phi(d_{ij})) \\
\mathbf{x}_i^{(\ell)} &= \mathbf{x}_i^{(\ell)} + \text{FFN}(\text{LN}(\mathbf{x}_i^{(\ell)}))
\end{align}
where $\text{EdgeAttn}$ is multi-head attention with edge bias $\phi(d_{ij})$ added to attention logits (Eq.~\ref{eq:edge_attn}), and:
\begin{equation}
\text{FFN}(\mathbf{x}) = \mathbf{W}_2 \text{Dropout}(\text{GELU}(\mathbf{W}_1 \mathbf{x} + \mathbf{b}_1), p=0.3) + \mathbf{b}_2
\end{equation}
with $\mathbf{W}_1 \in \mathbb{R}^{1536 \times 384}$, $\mathbf{W}_2 \in \mathbb{R}^{384 \times 1536}$ (expansion factor 4). Each layer contains $\sim$1.78M parameters; the 4-layer encoder totals 7,680,792 parameters.

\textbf{Output.} Final contextualized embeddings: $\mathbf{h}_i = \text{LN}(\mathbf{x}_i^{(L)})$.

\subsection{Node Classifier}

The node classifier applies a two-layer MLP to contextualized embeddings $\mathbf{h}_i \in \mathbb{R}^{384}$:
\begin{equation}
    \hat{\mathbf{y}}_i = \mathbf{W}_2 \text{Dropout}(\text{ReLU}(\mathbf{W}_1 \mathbf{h}_i + \mathbf{b}_1), p=0.1) + \mathbf{b}_2
\end{equation}
where $\mathbf{W}_1 \in \mathbb{R}^{256 \times 384}$, $\mathbf{W}_2 \in \mathbb{R}^{12 \times 256}$, and $\hat{\mathbf{y}}_i \in \mathbb{R}^{12}$ contains class logits. The architecture has 101,644 parameters.

\textbf{Hungarian Matching.} Given $N$ predictions and $M$ ground truth objects, we construct a cost matrix $\mathcal{C} \in \mathbb{R}^{N \times M}$ where:
\begin{equation}
\mathcal{C}(i, j) = \begin{cases}
1 - \text{softmax}(\hat{\mathbf{y}}_i)[c_j] & \text{if } c_j \in \mathcal{V}(s_i) \\
10^6 & \text{otherwise}
\end{cases}
\end{equation}
with $c_j$ the ground truth class, $s_i$ the SAM3 category of prediction $i$, and $\mathcal{V}(s)$ the compatibility mapping:
\begin{align}
\mathcal{V}(\text{person}) &= \{0, 2, 3, 4, 5, 6, 7\} \quad \text{(Patient, human\_0--5)} \\
\mathcal{V}(\text{table}) &= \{9, 10, 11\} \quad \text{(instrument/operating/secondary table)} \\
\mathcal{V}(\text{anesthesia}) &= \{1\} \quad \text{(anesthesia equipment)} \\
\mathcal{V}(\text{tool}) &= \{8\} \quad \text{(instrument)} \\
\mathcal{V}(\text{hand}) &= \emptyset
\end{align}
We solve $\sigma^* = \arg\min_{\sigma \in \mathfrak{S}_K} \sum_{i=1}^{K} \mathcal{C}(i, \sigma(i))$ via the Hungarian algorithm and filter pairs with $\mathcal{C} \geq 10^5$.

\subsection{Edge Classifier}

For each ordered node pair $(i,j)$, we construct pairwise features by concatenating:
\begin{equation}
\mathbf{e}_{ij} = [\text{LN}(\mathbf{h}_i, \mathbf{h}_j, \mathbf{h}_i - \mathbf{h}_j), \text{LN}(\mathbf{p}_i, \mathbf{p}_j), \mathbf{c}_i, \mathbf{c}_j, \mathbf{g}^{\text{geo}}_{ij}]
\end{equation}
where LN denotes layer normalization, $\mathbf{p}_i = \text{softmax}(\hat{\mathbf{y}}_i)\text{.detach}()$ are node class priors, $\mathbf{c}_i \in \mathbb{R}^5$ are SAM3 category hints, and $\mathbf{g}^{\text{geo}}_{ij} \in \mathbb{R}^{128}$ is the projected pairwise geometry.

\textbf{Pairwise Geometry.} Raw 11-dimensional geometric features are computed from cached data:
\begin{equation}
\mathbf{g}_{ij} = \left[d_{ij}, \hat{\mathbf{r}}_{ij}, \Delta z_{ij}, \log\left(\frac{\mathbf{s}_j}{\mathbf{s}_i} + 1\right), \log\left(\frac{V_j}{V_i} + 1\right), \text{IoU}_{3D}(i,j), \frac{d_{ij}}{\bar{s}_{ij}}\right]
\end{equation}
where $d_{ij} = \|\mathbf{p}_j - \mathbf{p}_i\|_2$, $\hat{\mathbf{r}}_{ij} = (\mathbf{p}_j - \mathbf{p}_i)/d_{ij}$ (3D), $\Delta z_{ij} = p_{j,z} - p_{i,z}$, $\mathbf{s}_i$ are bounding box dimensions (3D), $V_i$ is volume, and $\bar{s}_{ij} = (\|\mathbf{s}_i\| + \|\mathbf{s}_j\|)/2$. This is projected via a 2-layer MLP: $\mathbf{g}^{\text{geo}}_{ij} = \text{LN}(\mathbf{W}_2 \text{ReLU}(\text{LN}(\mathbf{W}_1 \mathbf{g}_{ij})))$ with $\mathbf{W}_1 \in \mathbb{R}^{128 \times 11}$, $\mathbf{W}_2 \in \mathbb{R}^{128 \times 128}$ (18,560 parameters).

\textbf{Edge Transformer.} The $N^2$ edge features are flattened to a sequence, projected to dimension 512, and processed through a 2-layer transformer encoder (4 heads, FFN expansion 4$\times$, 6.3M parameters):
\begin{equation}
\mathbf{E}' = \text{TransformerEncoder}(\mathbf{W}_{\text{proj}} \mathbf{E})
\end{equation}
After reshaping to $[B, N, N, 512]$, relation logits are computed: $\hat{\mathbf{r}}_{ij} = \mathbf{W}_{\text{out}} \text{LN}(\mathbf{E}'_{ij})$ with $\mathbf{W}_{\text{out}} \in \mathbb{R}^{15 \times 512}$ (8,719 parameters). Total edge head: 6.99M parameters.

\subsection{Loss}

The total loss combines node, edge, and proxy terms with weights $\lambda_{\text{node}}=1.0$, $\lambda_{\text{edge}}=2.0$, $\lambda_{\text{proxy}}=1.0$:
\begin{equation}
\mathcal{L} = \lambda_{\text{node}} \mathcal{L}_{\text{node}} + \lambda_{\text{edge}} \mathcal{L}_{\text{edge}} + \lambda_{\text{proxy}} \mathcal{L}_{\text{proxy}}
\end{equation}

\textbf{Node Loss.} Cross-entropy over matched pairs with uniform weighting:
\begin{equation}
\mathcal{L}_{\text{node}} = \frac{1}{B'} \sum_{b=1}^{B'} \frac{1}{K_b} \sum_{k=1}^{K_b} -\log \text{softmax}(\hat{\mathbf{y}}^b_{\sigma^*(k)})[c^b_k]
\end{equation}
where $B'$ is the number of batch elements with matches, $K_b$ is the number of matched nodes, and $\sigma^*$ are the Hungarian-matched indices.

\textbf{Edge Loss.} Inverse frequency-weighted cross-entropy over matched node pairs:
\begin{equation}
\mathcal{L}_{\text{edge}} = \frac{1}{B'} \sum_{b=1}^{B'} \frac{1}{K_b^2} \sum_{(i,j) \in \mathcal{E}_b} -w_{r_{ij}^*} \log \text{softmax}(\hat{\mathbf{r}}_{ij})[r_{ij}^*]
\end{equation}
where $\mathcal{E}_b$ contains all $K_b^2$ matched pairs (including diagonal), $r_{ij}^*$ is the ground truth relation, and class weights are:
\begin{equation}
w_r = \begin{cases}
1 / n_r & r = 0, \ldots, 13 \\
0.0001 & r = 14 \text{ (none)}
\end{cases}
\end{equation}
with $n_r$ the training set count of relation $r$.

\textbf{Proxy Loss.} Frame-level multi-label BCE for auxiliary supervision of rare relations:
\begin{equation}
\mathcal{L}_{\text{proxy}} = \frac{1}{14} \sum_{r=1}^{14} -\left[y_r \log \sigma(\hat{y}_r) + \alpha_r (1-y_r) \log(1 - \sigma(\hat{y}_r))\right]
\end{equation}
where $\hat{y}_r$ is the frame-level logit from a 2-layer MLP (76,815 parameters) applied to masked-mean pooled node embeddings, $y_r \in \{0,1\}$ indicates if relation $r$ exists in the frame, and $\alpha_r = n_{\text{neg},r} / n_{\text{pos},r}$ is the pos\_weight balancing factor. The \texttt{none} relation (index 14) is excluded from proxy supervision.

\section{Control Study: Prompt Selectivity}
\label{sec:control}

To verify that the +10 F1 improvement from adding unsupervised hand nodes reflects genuinely disambiguating context rather than an artifact of increased graph density, we conducted a control study with three additional prompts that are irrelevant to the surgical task: \textit{knee} (redundant with patient entity), \textit{door} (OR infrastructure), and \textit{shoe} (detected on clinicians' feet).

Table~\ref{tab:control_detections} shows the detection statistics for these control prompts. All prompts produce substantial numbers of detections (door and shoe each contribute more nodes per frame than hand), yet none improve F1 (see Table~\ref{tab:control}). This confirms that the model selectively leverages informative context: hands provide spatially and semantically disambiguating information for action-oriented relations, while irrelevant entities are effectively ignored by the attention mechanism.

\begin{table}[t]
\centering
\caption{Detection statistics for control study prompts across all 6,734 frames. Despite contributing comparable or more nodes than hands, irrelevant prompts provide no F1 improvement.}
\label{tab:control_detections}
\small
\begin{tabular}{@{}lccc@{}}
\toprule
Category & Total Detections & Avg/Frame & F1 Impact \\
\midrule
Hand (baseline) & 31,673 & 4.7 & +10 \\
\midrule
Knee (irrelevant) & 5,916 & 0.9 & 0 \\
Door (irrelevant) & 29,905 & 4.4 & 0 \\
Shoe (irrelevant) & 32,264 & 4.8 & 0 \\
\bottomrule
\end{tabular}
\end{table}

\end{document}